\documentclass{article}

\usepackage[a4paper,top=2.54cm,bottom=2.54cm,left=3.17cm,right=3.17cm,%
            includehead,includefoot]{geometry}
\usepackage{amsmath,amssymb,amsfonts,amsthm}
\usepackage{graphicx}
\usepackage{subfigure}
\usepackage{float}
\usepackage{xcolor}
\usepackage[numbers,square,sort&compress]{natbib}
\usepackage{hyperref}
\hypersetup{colorlinks,citecolor=blue,linkcolor=blue,breaklinks=true}
\usepackage{booktabs}
\usepackage{colortbl}
\usepackage{caption}
\usepackage{enumitem}
\usepackage{algorithm}
\usepackage{algpseudocode}
\usepackage{array}
\usepackage{bbding}
\usepackage{fancyhdr} 
\usepackage{fancyvrb} 
\usepackage{longtable}
\usepackage{listings}
\usepackage{rotating,rotfloat} 
\usepackage{yhmath} 
\usepackage{tikz}
\usetikzlibrary{3d, positioning, shapes.geometric, arrows.meta, calc}
\usepackage{stmaryrd}
\usepackage{multirow, bigstrut}

\usepackage{graphicx}  

\usepackage{fancyhdr}
\allowdisplaybreaks

\newtheorem{remark}{Remark}

\newcommand{\R}{\mathbb{R}}

\newcommand{\bbm}{\begin{bmatrix}}
\newcommand{\ebm}{\end{bmatrix}}

\title{A Hybrid Kernel-Free Boundary Integral Method with Operator Learning for Solving Parametric Partial Differential Equations In Complex Domains}

\author{Shuo Ling\thanks{These authors contributed equally to this work.} \\
    School of Mathematical Sciences, \\
    Shanghai Jiao Tong University, \\
    Shanghai 200240, P.R. China
    \and
    Liwei Tan\footnotemark[1] \\
    School of Mathematical Sciences, \\
    Shanghai Jiao Tong University, \\
    Shanghai 200240, P.R. China
    \and
    Wenjun Ying\thanks{Corresponding author. Email: \href{mailto:wying@sjtu.edu.cn}{wying@sjtu.edu.cn}} \\
    School of Mathematical Sciences, MOE-LSC and Institute of Natural Sciences, \\
    Shanghai Jiao Tong University, Minhang, \\
    Shanghai 200240, P.R. China.
}

\date{\today}

\begin{document}
\maketitle

\begin{abstract}
The Kernel-Free Boundary Integral (KFBI) method presents an iterative solution to boundary integral equations arising from elliptic partial differential equations (PDEs). This method effectively addresses elliptic PDEs on irregular domains, including the modified Helmholtz, Stokes, and elasticity equations. The rapid evolution of neural networks and deep learning has invigorated the exploration of numerical PDEs. An increasing interest is observed in deep learning approaches that seamlessly integrate mathematical principles for investigating numerical PDEs. We propose a hybrid KFBI method, integrating the foundational principles of the KFBI method with the capabilities of deep learning. This approach, within the framework of the boundary integral method, designs a network to approximate the solution operator for the corresponding integral equations by mapping the parameters, inhomogeneous terms and boundary information of PDEs to the boundary density functions, which can be regarded as the solution of the integral equations. The models are trained using data generated by the Cartesian grid-based KFBI algorithm, exhibiting robust generalization capabilities. It accurately predicts density functions across diverse boundary conditions and parameters within the same class of equations. Experimental results demonstrate that the trained model can directly infer the boundary density function with satisfactory precision, obviating the need for iterative steps in solving boundary integral equations. Furthermore, applying the inference results of the model as initial values for iterations is also reasonable; this approach can retain the inherent second-order accuracy of the KFBI method while accelerating the traditional KFBI approach by reducing about 50\% iterations.
\end{abstract}

\section{Introduction and Related Works}
Elliptic problems are widely applied in the fields of electrochemistry \cite{QIAN2021109908,DING2019108864}, electromagnetism \cite{Chai2023}, computational fluid dynamics \cite{Greengard1998318, Quartapelle1993NumericalSO}, shape optimisation problems \cite{ZHU2011752,gong2023} and other areas in science \cite{Chapko199747,ZHOU20061,CHENG2006616,SUN2014445}. Representative methods for solving specific elliptic problems numerically are the finite difference method \cite{zhou2006high, fedkiw1999non, leveque1994immersed, mccorquodale2001cartesian, berthelsen2004decomposed,johansen1998cartesian,twizell1996second,wiegmann2000explicit,hou2012numerical}, finite element method \cite{li2003newinterface,li1999fem,HUANG2017439,chu2010, wen2018finite, HOU2005411} , boundary integral method \cite{jwason1963,YING2004591,YING2006247,greengard_rokhlin_1997}. Some numerical methods based on deep learning \cite{E2018deepRitz,Raissi2019a,lulu2021deeponet,li2021fourier,Zang_2020,fan2023decoupling,HU2022111576,HE2022114358} are becoming popular in recent years. As a competitive approach to traditional and new methods, the Kernel-Free Boundary Integral (KFBI) method \cite{ying2007kernel,ying2014kernel,ying2013kernel} has shown its advantages .

The KFBI method is executed on a Cartesian grid for the resolution of general elliptic PDEs within domains of irregular shape with smooth perimeters. The KFBI method iteratively addresses the boundary integral equations and maintains symmetry and positive definiteness in the resultant discrete systems. This preservation enables the employment of effective solution strategies, including FFT-based or geometric multigrid solvers. Originating in the boundary integral method, the KFBI method not only retains the favourable conditioning of the boundary integral equation but also obviates the direct computation of Green's function, which is notably complex in irregular domains \cite{xie2019fourth, ying2013kernel}. Consequently, its effectiveness is particularly notable in overcoming computational mathematics challenges. In recent years, the KFBI method has been extensively applied\cite{xie2019fourth,xie2021cartesian,ZHAO2023116163,dong2023kernelfree,zhou2023adi}.


Deep learning methodologies have been acknowledged for their transformative potential in scientific research, offering accelerated solutions that approximate or surpass traditional methods in some specific scenarios \cite{Jiang2020a, Raissi2019, Greenfeld2019, Kochkov2021a}. Deep neural networks (DNNs) have been increasingly utilized for solving PDEs, circumventing the explicit discretization requirement, and learning mappings in specific spaces favorable for discovering solutions to PDEs. Simultaneously, DNN-based methods demonstrate efficacy in mitigating the curse of dimensionality and prove beneficial in addressing certain inverse problems. Various neural network architectures, loss functions, and activation functions have been explored for this purpose. For instance, the deep Galerkin method (DGM) \cite{sirignano2018dgm} and physics-informed neural networks (PINNs) \cite{raissi2019physics} employ equation residuals as loss functions within a general framework for PDE resolution. These networks are refined through stochastic gradient descent, applying spatial point random sampling within the domain. Implement of conditions is achieved by network integration \cite{berg2018unified} or loss penalization, with the latter relying on penalty coefficients as hyper-parameters. However, the fine-tuning of these coefficients is a complex process necessitating further methodical investigation. A recently proposed boundary integral network (BINet) \cite{CSIAM-AM-4-275} presented a convolution representation of the solutions to elliptic PDEs using Green's functions. This approach was subsequently extended to a method for discovering general Green's functions, which can be acquired through training neural networks \cite{lin2023bi}.

In the context of operator learning, classical neural networks, which are confined to mapping between finite-dimensional spaces, face limitations in learning discretization-specific solutions on two-dimensional or higher-dimensional grids. This necessitates the development of mesh-invariant neural networks and other DNN-based methods who can play the role of reducing the dimension of operator learning. Recent studies have introduced the concept of learning mesh-free, infinite-dimensional operators using neural networks \cite{li2021fourier,lulu2021deeponet,Bhattacharya2021,Nelsen2021,Patel2021}. These neural operators, capable of resolving a class of PDEs rather than specific instances, allow evaluations at arbitrary temporal and spatial points. As an example, the Fourier Neural Operator (FNO) is proposed in \cite{li2021fourier}, utilizing Fourier transformation for network architecture design. For DeepONet proposed in \cite{lulu2021deeponet}, a network structure comprising branch and trunk nets is introduced, addressing PDE parameters and spatial coordinates, respectively. Additionally, Deep Green \cite{Gin2021} and MOD-net \cite{CiCP-32-299} employ neural networks to approximate Green's function, effectively representing the solution operator for nonlinear PDEs by mapping source terms or boundary values to solutions.

To a certain extent, the previously mentioned class of operator learning methods appears to lack precision assurances, with some methods exhibiting less satisfactory accuracy. This deficiency can be attributed to inadequate utilization of mathematical prior knowledge. In this work, we introduces a novel approach called hybrid kernel-free boundary integral (hybrid KFBI) method that integrates operator learning with the KFBI method. The method solves equations within the framework of boundary integral methods, featuring DNNs designed to approximate the solution operators of the corresponding boundary integral equations, which maps from the parameters, inhomogeneous terms and boundary information of PDEs to the boundary density functions. The hybrid KFBI Method has advantages including:

\textbf{High Data Quality:} KFBI is fundamentally a boundary integral method that transforms two-dimensional problems into one-dimensional boundary problems, achieving dimensionality reduction in the model. Furthermore, due to its high precision, the KFBI method can produce high-quality data for training process of hybrid KFBI method. 

\textbf{High Precision}: The trained model can directly predict density functions (instead of solving the boundary integral equations iteratively) for computing solutions for the original PDEs, significantly reducing the solution time for the KFBI method while maintaining high accuracy (relative error in the order of 1E-3 or 1E-4). It can also be employed as an initial value, substantially decreasing the number of iterations required without compromising the accuracy relative to the KFBI method.

\textbf{Strong Generalizability:} Based on our meticulously designed network architecture and input-output methodology, each trained model is applicable to a broad range of equations. For instance, when elastic equations are considered, the different inhomogeneous terms, boundary conditions, and physical parameters can be  inputted into the same model, which demonstrates robust generalization capabilities.

\textbf{Model Dimensionality Reduction:} The boundary density functions are defined on the boundary whose dimension is less by 1 than the original computational domain, which is helpful for operator learning. Lower dimension leads to fewer sampling points which will reduce the computational cost.

\textbf{High Integration Capability:} Our approach possesses the ability to integrate with certain other methods, such as those reliant on high-performance computing using GPUs, thereby achieving a more efficient solution efficiency.

The structure of this paper is methodically organized as follows. Initially, the fundamental theory underlying the boundary integral (BI) method and the KFBI method is introduced in Section $\ref{BIM}$. Following this, Section $\ref{Hybrid}$ elaborates on various aspects of the hybrid KFBI method in detail, which integrates operator learning techniques to enhance its efficacy and versatility. Subsequently, Section $\ref{Numerical Experiments}$ is dedicated to presenting the numerical results obtained. The concluding Section \ref{Conclusion} engages in a comprehensive discussion regarding the merits, limitations, and prospects of the hybrid KFBI method, encapsulating the essence of this research.

\section{Preliminaries} \label{BIM}
In this section, we will expound upon a segment of knowledge pertaining to BI methods, serving as the foundation for both the KFBI method and the hybrid KFBI approach. Let $\Omega$  be a bounded, two-dimensional domain of irregular and complex structure, with a boundary $\Gamma = \partial \Omega$ possessing at least $C^2$ continuity. The function $u(\mathbf{x}) = u(x, y)$, where $\mathbf{x} \in \mathbf{R}^{2}$, remains undetermined (analogous considerations apply to $\mathbf{R}^d$ for $d > 2$). The foundational principles of the BI method are discussed for the Dirichlet boundary condition. The case with the Neumann condition is addressed in Appendix \ref{appen::neumann::condition}.

\begin{figure}[ht]
    \centering
    \includegraphics[width = 0.8\linewidth]{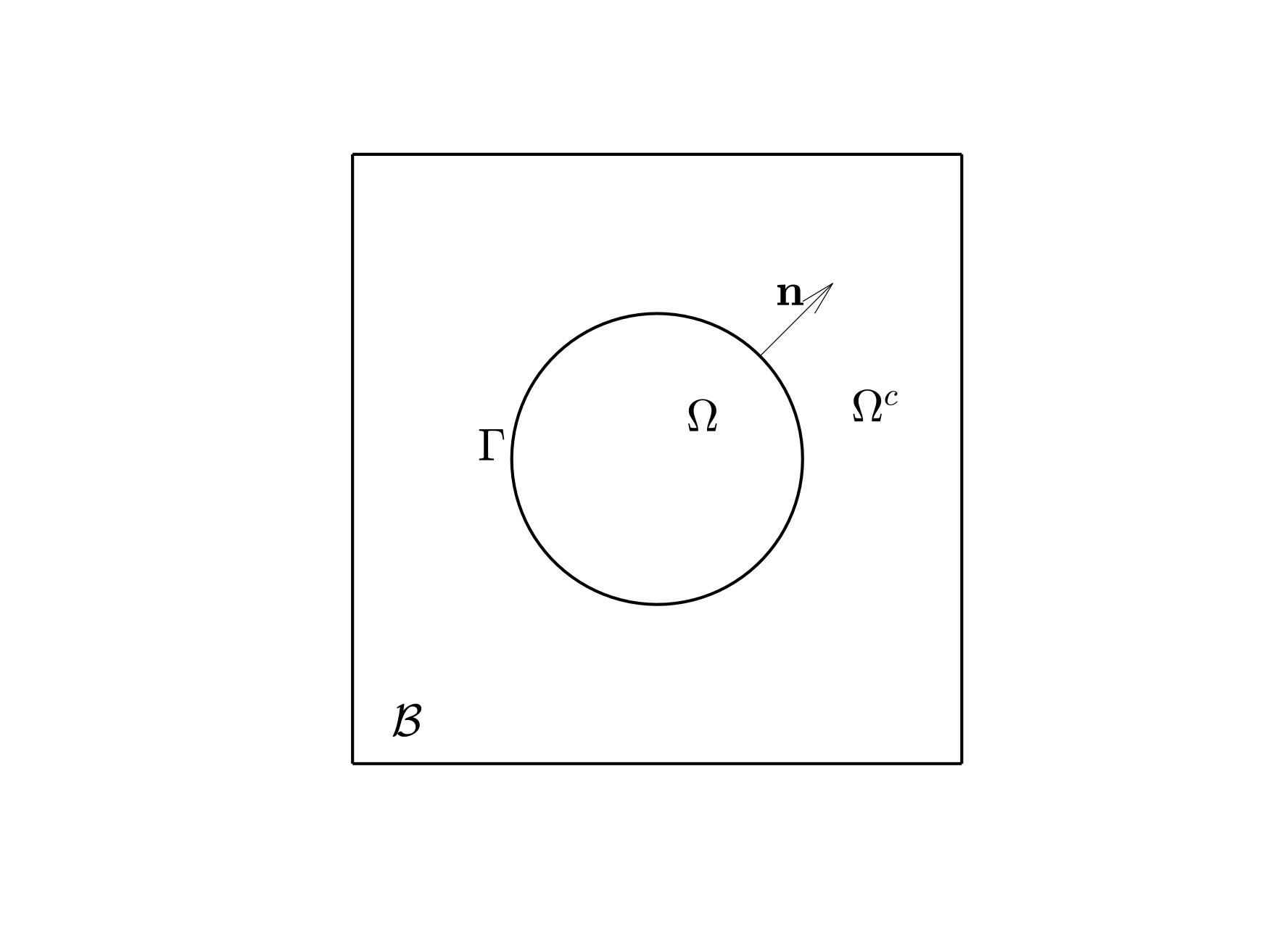}
    \caption{BI method and KFBI method computation domain}
    \label{kfbi domain}
\end{figure}

As shown in Fig.\,$\ref{kfbi domain}$, to solve the boundary value problems on domain $\Omega$ by the BI method, we first embed the irregular domain $\Omega$ into a larger rectangle domain $\mathcal{B}$ and denote by $\Omega^{c} = \mathcal{B} \backslash \Omega$ the complement of the domain $\Omega$ in $\mathcal{B}$. 

Suppose $\mathcal{L}$ is an elliptic operator, consider the elliptic equation 
\begin{equation}
   (\mathcal{L} u)(\mathbf{x}) =f(\mathbf{x}), \quad \text { in } \Omega,
    \label{one_GPU:PDE}
\end{equation}
where $\mathcal{L} = \nabla \cdot \sigma(\mathbf{x}) \nabla-\kappa(\mathbf{x})$ with the diffusion tensor $\sigma$ and the reaction coefficient $\kappa \geq 0$ which are at least continuously differentiable over the regular domain $\mathcal{B}$ \cite{ying2007kernel}. Subject to pure Dirichlet boundary condition 
\begin{equation}
    u(\mathbf{x}) = g_{D}(\mathbf{x}), \quad \text{ on }\Gamma,
    \label{one_GPU:PDE_D_BC}
\end{equation}
where $f(\mathbf{x})\ \mbox{and}\ g_D(\mathbf{x})$  are known functions of $\mathbf{x}=(x, y)$ with sufficient smoothness for solving the corresponding equation.

According to the standard BI method \cite{aliabadi2011boundary,yu2002natural}, let $G(\mathbf{x}, \mathbf{y})$ be Green's function on the rectangle $\mathcal{B}$ associated with the elliptic PDE $\eqref{one_GPU:PDE}$, which satisfies for any $\mathbf{y} \in \mathcal{B}$,
\begin{align}
\mathcal{L} G(\mathbf{x}, \mathbf{y}) &= \delta(\mathbf{x}-\mathbf{y}), \quad \mathbf{x} \in \mathcal{B}, \\
G(\mathbf{x}, \mathbf{y}) &=0, \quad \mathbf{x} \in \partial \mathcal{B},
\end{align}
where $\delta(\mathbf{x} - \mathbf{y})$ is the Dirac delta function. Let $\mathbf{n}_{\mathbf{y}}$ be the unit outward normal vector at point $\mathbf{y}\in \Gamma$, $\phi$ be a function defined on the boundary $\Gamma$ and $F$ be a function defined on $\Omega$, we firstly define the double layer boundary integral and volume integral by
\begin{align}
(W \phi)(\mathbf{x}) := \int_{\Gamma} \frac{\partial G(\mathbf{y}, \mathbf{x})}{\partial \mathbf{n}_{\mathbf{y}}} \phi(\mathbf{y}) d s_{\mathbf{y}}, \quad & \text { for } \mathbf{x} \in \Omega \cup \Omega^{c},\label{one_GPU:double} \\
(YF)(\mathbf{x}) := \int_{\Omega} G(\mathbf{y}, \mathbf{x})F(\mathbf{y}) d\mathbf{y}, \quad & \text{ for } ~\mathbf{x} \in \Omega.
    \label{one_GPU:volume}
\end{align}


Thanks to the symbols and properties of the involved potential and volume integral, the boundary integral equation(BIE) about the density function $\varphi$ for $\eqref{one_GPU:PDE}$-$\eqref{one_GPU:PDE_D_BC}$ can be reformulated as a Fredholm boundary integral equation of the second kind \cite{kress1989linear,hsiao2008boundary}
\begin{equation}
    \frac{1}{2}\varphi (\mathbf{x}) + (W\varphi)(\mathbf{x}) + (Yf)(\mathbf{x}) = g_{D}(\mathbf{x}), \quad \mathbf{x} \in \Gamma. \label{one_GPU:second_fredholm}
    %
\end{equation}

The solution $u(\mathbf{x})$ to the Dirichlet BVP $\eqref{one_GPU:PDE}$-$\eqref{one_GPU:PDE_D_BC}$ is given by 
\begin{equation}
    u(\mathbf{x}) = (W\varphi)(\mathbf{x}) + (Yf)(\mathbf{x}), \quad \mathbf{x} \in \Omega, \label{One_GPU:fredholm_dirichlet} \\
\end{equation}
where $\varphi$ is the solution of equation \eqref{one_GPU:second_fredholm}.

From $\eqref{one_GPU:second_fredholm}$, it is convenient to denote 
\begin{align}
    \widetilde{W}(\varphi)(\mathbf{x}) & := (\frac{1}{2} \mathcal{I} + W)(\varphi)(\mathbf{x}), \quad \mathbf{x} \in \Gamma,
\end{align}
where $\mathcal{I}$ is the identity operator.

In the KFBI method, the boundary integral equation $\eqref{one_GPU:second_fredholm}$ can be solved by the Richardson iteration numerically. Given any initial guess $\varphi_{0}(\mathbf{x})$, for $k = 0, 1, 2,  \cdots$, perform the following steps until convergence (within a predefined tolerance)
\begin{align}
    u_{k}^{+}(\mathbf{x}_{m}) = \widetilde{W}\varphi_{k}(\mathbf{x}_{m}), & \quad m = 1, 2, \cdots, M, \label{one_GPU:richardson1} 
\end{align}
\begin{align}
     \varphi_{k+1}(\mathbf{x}_{m}) = \varphi_{k}(\mathbf{x}_{m}) + 2 \gamma [\widetilde{g_{D}}(\mathbf{x}_{m}) - u_{k}^{+}(\mathbf{x}_{m})], & \quad m = 1, 2, \cdots, M, \label{one_GPU:richardson2} 
\end{align}
where $\{\mathbf{x}_i\}_{i = 1}^M$ are discrete points on boundary $\Gamma$ and $\widetilde{g_{D}}(\mathbf{x}) := g_{D}(\mathbf{x}) - (Yf)(\mathbf{x})$ need only calculate once before Richardson iteration. Note that equation \eqref{one_GPU:richardson1} is from potential theory about double potential \cite{jwason1963} instead of an arbitrary definition. It can be shown by considering the spectral radius of $\widetilde{W}$ that the Richardson iteration is convergent if $\gamma \in (0, 1)$.  The superscript `+' in the BIE means one-sided limit from the domain $\Omega$. More specifically, let $w(\mathbf{x})$ be an arbitrary piecewise smooth function with discontinuities only existing at the interface $\Gamma$. We denote
 \begin{equation}
     w^{+}(\mathbf{x}) = \lim_{\mathbf{z} \to \mathbf{x}, \mathbf{z} \in \Omega} w(\mathbf{z}).
 \end{equation}

Similarly, the restriction of $w(\mathbf{x})$ in $\Omega^{c}$, $w^{-}(\mathbf{x})$ is defined as 
\begin{equation}
    w^{-}(\mathbf{x}) = \lim_{\mathbf{z} \to \mathbf{x}, \mathbf{z} \in \Omega^{c}} w(\mathbf{z}).
\end{equation}

Once the unknown density function $\varphi(\mathbf{x})$ is obtained when the iteration $\eqref{one_GPU:richardson2}$ converges. The unknown function $u(\mathbf{x})$ can be calculated according to the formula $\eqref{One_GPU:fredholm_dirichlet}$.

The remaining task involves determining an efficient approach for computing volume integrals $Yf$ and boundary integrals $W \varphi$ to achieve the equations \eqref{one_GPU:richardson1} during iteration. The KFBI method offers a methodology that eliminates the need for an explicit expression of the Green's function. This involves transforming  computation of volume and boundary integrals into  solution of corresponding interface PDEs whose details are shown in Appendix \ref{introduce_kfbi}.

\section{Methodology for Hybrid KFBI Method}
\label{Hybrid}
This section introduces the hybrid KFBI method, which combines traditional KFBI methods detailed in Appendix \ref{introduce_kfbi} with deep learning techniques, emphasizing efficient solutions for multiple instances of the same type of PDEs. A specialized class of networks designed to learn the solution operator for boundary integral equations will be integrated into the KFBI framework for PDE resolution. Attention is restricted to PDEs subject to Dirichlet boundary conditions. However, it is noted that Neumann boundary conditions are addressed within a similar framework, which will not be extensively discussed here.

\subsection{Solution Operator for Integral Equations}
In the KFBI method, the boundary integral equation \eqref{one_GPU:second_fredholm} is resolved iteratively, utilizing methods like Richardson iteration. Given that equation \eqref{one_GPU:second_fredholm} remains non-singular for a general symmetric positive definite diffusion tensor $\sigma$ and non-negative reaction coefficient $k$ \cite{ying2007kernel}, the solution operator $\mathcal{S}_{\mathcal{L}, \Omega}$ can be defined as mapping the modified boundary value $\widetilde{g_D}$ to the density $\varphi$. The encapsulation of the right-hand side \(f\) of equation \eqref{one_GPU:PDE} and the boundary condition $g_D$ into $\widetilde{g_D}$ enables the learning of $\mathcal{S}_{\mathcal{L}, \Omega}$, thereby facilitating the resolution of a wide range of equations with variable \(f\) and assorted boundary conditions.

\begin{remark} \label{rmk_linearity}
For fixed elliptic operator $\mathcal{L}$ and domain $\Omega$ satisfying the requirement of the KFBI method, the operator $\mathcal{S}_{\mathcal{L}, \Omega}$ between the function spaces defined above is linear.
\end{remark}



\subsection{Operator Learning} \label{Operator Learning}
A natural and appealing wish is to obtain the explicit or computationally convenient form of the operator $\mathcal{S}_{\mathcal{L}, \Omega}$, which was nearly impossible in the past. With advanced deep learning and related technologies, models based on deep neural networks now offer the potential to realize this vision. In other words, we can use neural networks to approximate this operator and aim to obtain the best network parameters through training on data for the most effective approximation.

We provide a detailed exposition of the methodology tailored for $\Omega \in \R^2$, noting that analogous approaches apply to higher-dimensional contexts. Recall that in the standard KFBI method, we can discretize the larger rectangle $\mathcal{B}$ using a rectangular grid with cell numbers $I \times J$, for which $\mathcal{B} \supset \Omega$ and $\partial\Omega\cap\partial\mathcal{B}=\emptyset$. The discretization of the density $\varphi$ is achieved through periodic cubic splines, facilitating the straightforward and efficient computation of density derivatives. More precisely, we presuppose the existence of $M$ quasi-uniformly spaced nodes on the boundary $\partial \Omega$. Note the meanings of \(I\), \(J\), and \(M\) will be consistently employed throughout the subsequent sections of this paper. To better integrate with the standard KFBI method, a natural and effective approach for operator learning is to set the input and output of the neural network to be the values of $\widetilde{g_D}$ and $\varphi$ at the $M$ points on the closed curve $\partial \Omega$, respectively, when using network $\mathcal{N}_{\mathcal{L}, \Omega}$ to approximate the operator $\mathcal{S}_{\mathcal{L}, \Omega}$. In addition, we facilitate the utilization of the standard KFBI method for the generation of multiple datasets (which will be elaborated upon in detail in subsection \ref{Strategies for Generating Training Data}), expediently employed in the training of the neural network.

If the neural network $\mathcal{N}_{\mathcal{L}, \Omega}$ can be successfully trained and possesses high approximation accuracy and strong generalization, it will significantly aid in solving equations $\mathcal{L} u = f$ in the fixed domain \(\Omega\) with free inhomogeneous term $f$ and Dirichlet-type boundary conditions $u|_{\partial \Omega} = g^D$ by KFBI method (the role of the model, or its specific application, will be discussed in detail in next section \ref{Hybrid KFBI Method Based on Trained Models}). The network $\mathcal{N}_{\mathcal{L}, \Omega}$ has already been demonstrated to yield satisfactory results in subsequent numerical examples. 

However, a more ambitious idea is to explore incorporating entire or partial boundary information yielded by boundary curves into the network, enabling it to operate beyond the constraints of solving equations solely within a fixed region. Considering the difficulties posed by arbitrarily closed curves are incalculable, a more common approach is considering a class of parametric curves. Specifically, $\mathcal{S}_{\mathcal{L}}$ can be considered as a solution operator for equation $\frac{1}{2} \varphi+\mathcal{K} \varphi =  \widetilde{g_D}$ (caused by the fixed operator $\mathcal{L}$) on a class of $\partial \Omega$, which is mapping $(\partial \Omega, \widetilde{g_D})$ to the density $\varphi$. Thus, we aspire for a neural network $\mathcal{N}_{\mathcal{L},}$ that can simultaneously intake boundary position information of $\partial \Omega$ and values about the function $\widetilde{g_D}$, yielding output values about density $\varphi$. Denote by $\{ \mathbf{x_i} \} _{i = 1}^{M}$ the point set in the KFBI method which are used to discrete the boundary $\partial \Omega$. Recall that in the previous network $\mathcal{N}_{\mathcal{L}, \Omega}$, we configured the input of this network as $(\widetilde{g_D}(\mathbf{x_1}), \widetilde{g_D}(\mathbf{x_2}), ..., \widetilde{g_D}(\mathbf{x_M}))^T$ and the output as $(\varphi(\mathbf{x_1}), \varphi(\mathbf{x_2}), ..., \varphi(\mathbf{x_M}))^T$. Returning to the discussion about network $\mathcal{N}_{\mathcal{L}}$, the required input should be divided into two parts. One component consists of the values of the function $\widetilde{g_D}$, which remains consistent with the above, while the other component should incorporate spatial location information yielded by a class of boundary, such as ellipses with different major and minor axes. Here, we propose two methods for incorporating boundary position information into the network. One approach is to directly incorporate the positional coordinates into the network, that is, setting the network's input as $(\mathbf{x_1}, \mathbf{x_2}, ..., \mathbf{x_M})^T$ and $(\widetilde{g_D}(\mathbf{x_1}), \widetilde{g_D}(\mathbf{x_2}), ..., \widetilde{g_D}(\mathbf{x_M}))^T$, where $\mathbf{x_i} = (x_{i, 1}, x_{i, 2})^T$ is the coordinate for all $i = 1, 2, ..., N$. Another approach involves adding only the parameters of the boundary curve to the network's input. For instance, the ellipses with parameters of axial lengths $r_a, r_b$ can be defined as $\partial \Omega_{\{r_a, r_b\}} := \{(x, y): \frac{(x - x_0)^2}{r_a^2} + \frac{(y - y_0)^2}{r_b^2} = 1\}$, here $x_0$ and $y_0$ are constant numbers. Then the input of the network $\mathcal{N}_{\mathcal{L}}$ can be designed as $(r_a, r_b)^T$ and $(\widetilde{g_D}(\mathbf{x_1}), \widetilde{g_D}(\mathbf{x_2}), ..., \widetilde{g_D}(\mathbf{x_M}))^T$. While it is true that various methods can be suggested, these two approaches have been observed to be effective and computationally efficient in subsequent numerical experiments, especially the first approach that was able to include more information.

The parametric PDEs that we aim to investigate not only manifests in the ability to represent the domain boundaries parametrically but also occasionally entails solving equations induced by a parametric operator $\mathcal{L}$, such as the Helmholtz equations and screened Poisson equations. In this case, we need a neural network $\mathcal{N}_{\Omega}$ to approximate the solution operator $\mathcal{S}_{\Omega}$, which is defined as mapping $(\mathcal{L}, \widetilde{g_D})$ to the density $\varphi$ for a class of parameterized elliptic operator $\mathcal{L}$. Similar to the preceding discussion, we can incorporate the parameters of operator $\mathcal{L}$ into the network's input. Specifically, if the operator $\mathcal{L}$ has parameters $k_1, k_2, ..., k_n$, we can set the input of networks $\mathcal{N}_{\Omega}$ as $(k_1, k_2, ..., k_n)^T$ and $(\widetilde{g_D}(\mathbf{x_1}), \widetilde{g_D}(\mathbf{x_2}), ..., \widetilde{g_D}(\mathbf{x_M}))^T$. 

\begin{remark}
Our ultimate goal is to incorporate both the parameters of the equation operator $\mathcal{L}$ and the parameters of the boundary $\partial \Omega$ into the network $\mathcal{N}$, which is used to approximate the solution operator $\mathcal{S}$ of equation $\frac{1}{2} \varphi+\mathcal{K} \varphi =  \widetilde{g_D}$ caused by the parametric operator $\mathcal{L}$ and on a class of parametric $\partial \Omega$. Incorporating the operator's parameters $\mathcal{L}$ and the boundary $\partial \Omega$ parameters into the network does not pose new technical challenges. However, concerns arise regarding the difficulties in generating sufficient data and potential issues related to the increased size of network parameters that might result in slower inference speeds. The subsequent numerical experimentation section will elucidate the attempts we made, while this could become a significant topic for consideration in our future exploration.
\end{remark}

\subsubsection{Network Architectures}
\textbf{Networks without Parameter Component}\hspace{0.5em}  The specific structure of the network $\mathcal{N}_{\mathcal{L}, \Omega}$, which receives $(\widetilde{g_D}(\mathbf{x_1}), ..., \widetilde{g_D}(\mathbf{x_M}))^T$ as input and outputs $(\varphi(\mathbf{x_1}), ..., \varphi(\mathbf{x_M}))^T$, is outlined in this section. Note that the network must possess a linear structure by remark \ref{rmk_linearity}. A fully connected neural network, comprising only input and output layers without or with linear activation functions, is typically utilized. A network directly mapping input to output with $M^2$ parameters is considered suitable. However, for large \(M\) values (e.g., $M = 1024$ when discretizing the boundary \(\partial \Omega\) with 1024 points), the number of network parameters can be substantial. To address this, a multi-layered network structure is recommended for large \(M\) values. This involves mapping the \(M\)-dimensional input to a hidden layer of \(M // d\) dimensions (where $d$ is a constant such as 2, 4, or 8, and the operator `//' signifies floor division), then mapping to another hidden layer of \(M // d\) dimensions, before reaching the \(M\)-dimensional output. The total parameters in this network are calculated as \(2 \times M \times (M // d) +  (M // d)^2\), significantly less than \(M^2\) for an appropriately chosen $d$.

\textbf{Networks with Parameter Component}\hspace{0.5em} In this part, the parameter component is incorporated into the input of the network $\mathcal{N}_{\mathcal{L}}$, $\mathcal{N}_{\Omega}$ or $\mathcal{N}$. As outlined in the previous subsection, this input component may extra include positional coordinates of points on the domain boundary $(\mathbf{x_1}, \mathbf{x_2}, ..., \mathbf{x_M})^T$, boundary parameters such as $r_a, r_b$, or operator parameters like $k_1, k_2, ..., k_n$. It is noted that if the quantity of parameters is insufficient, an additional preprocessing layer is introduced. This layer, a fully connected network (with or without an activation function), maps a limited number of initial input parameters to a broader set. All parameters from PDEs (after the preprocessing if it is necessary) are collectively denoted as $p_1, p_2, ..., p_N$ for clarity in the network structure description. Each $p_i$ can be a scalar or a coordinate, with $N$ representing the total number of parameters, including those from any preprocessing layer. Precise determination of these details based on the input and output shapes is crucial when constructing a specific network.

\begin{figure}[ht]
    \centering
    \includegraphics[width = 0.98\linewidth]{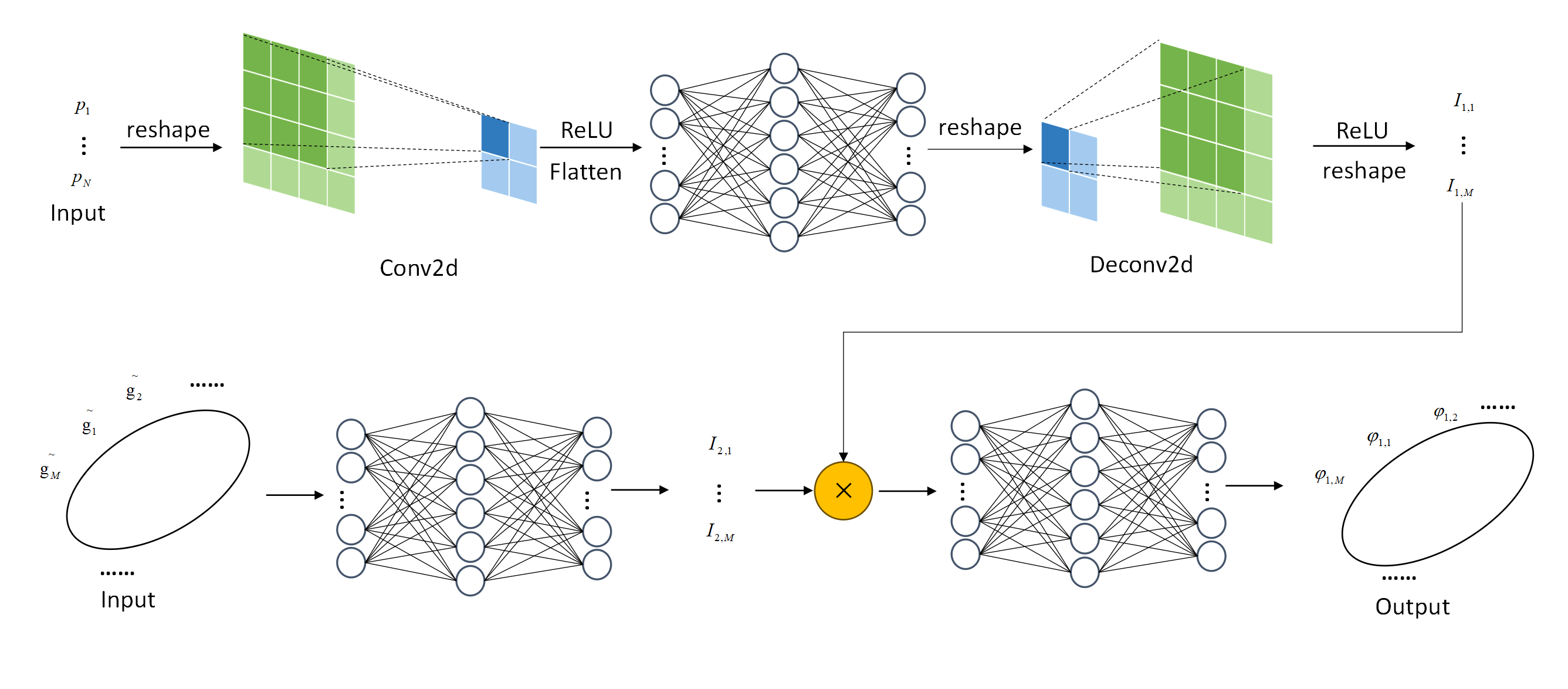}
    \caption{Network architecture schematic for networks with parameter component. In this graph, $(I_{1, 1}, I_{1, 2}, ..., I_{1, M})^{T} = I_1$ and $(I_{2, 1}, I_{2, 2}, ..., I_{2, M})^{T} = I_2$ are used to present the intermediate elements, $\widetilde{g}_i$ is used to present $\widetilde{g_D}(\mathbf{x_i})$ and $\varphi_i$ is used to present $\varphi(\mathbf{x_i})$ for all $i = 1, 2, ..., M$. The schematic diagram of fully connected networks do not represent the actual size of the network model in practical use, they are for illustrative purposes only and these three blocks do not share their parameters. Also note that the `Flatten' and `reshape' operations are performed to obtain the correct shapes of tensors, enabling them to be properly passed to the next network block, and some terms in this diagram have the same meanings as their corresponding terms in the \textbf{PyTorch} library, such as `Conv2d' represents 2D convolution and `ConvTransposed2d' represents 2D transposed convolution.} \label{fig:networks}
\end{figure}

A structure combining convolutional neural networks (without pooling layers) and fully connected networks is employed to handle the parameter component input $(p_1, p_2, ..., p_N)^T$. The aim is to produce an intermediate element, denoted as $I_1$, matching the dimensions of the input $(\widetilde{g_D}(\mathbf{x_1}), ..., \widetilde{g_D}(\mathbf{x_M}))^T$. Simultaneously, a linear transformation, devoid of nonlinear activation functions and maintaining the shape for $(\widetilde{g_D}(\mathbf{x_1}), ..., \widetilde{g_D}(\mathbf{x_M}))^T$, yields another intermediate element, denoted as $I_2$. Element-wise multiplication is then applied between these two intermediate elements. Following this, a linear structure similar to the previous section is applied to the post-multiplication result. Note that nonlinear operations, such as the ReLU activation function, acting only on the component of parameter input, this architectural design ensures input linearity for function values when input parameters remain constant while effectively utilizing the information in the parameter component of the input. The subsequent figure \ref{fig:networks} illustrates the specific network architecture and note that the network architecture is partly inspired by DeepONet \cite{lulu2021deeponet}.
\begin{remark}
The `parameter' mentioned in the preceding paragraph refers to the neural network's parameter input, denoted as $(p_1, p_2, ..., p_N)^T$, rather than the model parameters of the neural network itself. Please avoid any confusion between them.
\end{remark}

\subsubsection{Strategies for Generating Training Data} \label{Strategies for Generating Training Data}
Initially, with fixed $\mathcal{L}$ and $\Omega$, focus is placed on data generation for training the operator $\mathcal{N}_{\mathcal{L}, \Omega}$. For fixed operator $\mathcal{L}$ and domain $\Omega$, an exact solution to equation \eqref{one_GPU:PDE} can be constructed, and the KFBI method can be executed based on the inhomogeneous term $f$ and boundary condition $g_D$ derived from this exact solution. This process yields a pair $(\widetilde{g_D}, \varphi)$, assuming the KFBI method's minor error is negligible. Specifically, pairs $(\widetilde{g_D}(\mathbf{x_1}), ..., \widetilde{g_D}(\mathbf{x_M}))^T$ and $(\varphi(\mathbf{x_1}), ..., \varphi(\mathbf{x_M}))^T$ are obtained from each exact solution. To train the linear neural network $\mathcal{N}_{\mathcal{L}, \Omega}$ effectively, more than $M$ linearly independent vectors $\widetilde{g_D}$ should be added to the dataset. Consequently, constructing over $M$ exact solutions for equation \eqref{one_GPU:PDE} to obtain corresponding sets of $\widetilde{g_D}$ and $\varphi$ becomes necessary. Typically, $M$ can be chosen as a power of 2, like 128, 256, 512, etc. In general, to get a more generalized model, we will make the size of the data set much larger than $M$ although a training data set with $M$ linearly independent pairs is enough due to the linearity of real operators and the networks. For instance, several thousand pairs of $\widetilde{g_D}$ and $\varphi$ are enough to train the model when $M = 128$.

For training $\mathcal{N}_{\mathcal{L}}$, $\mathcal{N}_{\Omega}$ or $\mathcal{N}$, sufficient data sets are generated in the same manner for each parameter pair \((p_1, p_2, ..., p_N)\). These parameter pairs \((p_1, p_2, ..., p_N)\) are uniformly sampled within the given parameter space to ensure adequate data volume.
 
\subsubsection{Loss Function and Training Process}
For fixed $\mathcal{L}$ and $\Omega$, training the parameters $\Theta_{\mathcal{L}, \Omega}$ of operator $\mathcal{N}_{\mathcal{L}, \Omega}$ with dataset $\{ (\widetilde{g_D}_i, \varphi_i) \}_{i = 1}^n$ gained from the strategy mentioned above by using this trivial loss function:
\begin{equation} \label{eq3.2.3.1}
loss(\Theta_{\mathcal{L}, \Omega}) = \sum_{i = 1}^n || \mathcal{N}_{\mathcal{L}, \Omega}(\widetilde{g_D}_i; \Theta_{\mathcal{L}, \Omega}) - \varphi_i ||_2^2,
\end{equation}
where $\widetilde{g_D}_i = (\widetilde{g_D}_i(\mathbf{x_1}), \widetilde{g_D}_i(\mathbf{x_2}), ..., \widetilde{g_D}_i(\mathbf{x_M}))^T$, $\varphi_i = (\varphi_i(\mathbf{x_1}), \varphi_i(\mathbf{x_2}), ..., \varphi_i(\mathbf{x_M}))^T$ are the $i$-th pair of data and $||\cdot||_2$ is used to represent the $2$-norm of vectors. For training $\mathcal{N}_{\mathcal{L}}$, $\mathcal{N}_{\Omega}$ or $\mathcal{N}$, similar loss functions are utilized.

The dataset derived from the strategy above is divided into two parts for training: an $80\%$ training set and a $20\%$ test set through a random split. This division aims to mitigate overfitting and underfitting risks. The training process uses the \textbf{PyTorch} deep learning framework, with \textbf{Adam} as the optimizer. An initial learning rate of $0.001$ is set, and a learning rate scheduler with reduced rate $0.6$ is incorporated to dynamically adjust the learning rate if the loss exceeds 1500 times without a decrease. When no further reduction in the loss function is observed, typically after not exceeding 500,000 or 800,000 epochs, the training terminates, at which point the value of the loss function is generally within the magnitude of 1E-5 in our experiments. This training is executed on a single \textbf{NVIDIA GeForce RTX 3080} graphics card utilizing the \textbf{CUDA} platform.

\subsection{Hybrid KFBI Method Based on Trained Models} \label{Hybrid KFBI Method Based on Trained Models}
This section introduces the significant role offered by the trained neural networks $\mathcal{N}_{\mathcal{L}, \Omega}$ (or $\mathcal{N}_{\mathcal{L}}$, $\mathcal{N}_{\Omega}$, $\mathcal{N}$). The networks are designed to infer the density function $\varphi$. Consequently, the inference results from these neural networks can be directly utilized as density functions $\varphi$, which are then integrated into the formula \eqref{One_GPU:fredholm_dirichlet}. Solving the corresponding interface problem \eqref{one_GPU:interface} yields the solution to the PDE \eqref{one_GPU:PDE}. Alternatively, the output $\varphi$ from the neural networks can be used as the initial value in the iterative KFBI method, reducing the number of iterations and saving time. These approaches are subsequently referred to as Strategies 1 and 2, respectively. Regarding the utilization of neural network inference results as initial values, there exists lots of related works, each demonstrating commendable efficacy within their respective applications, such as \cite{luna2021accelerating}.

\begin{remark}
\begin{itemize}
\item The two methods above have been observed to significantly reduce iteration count and time compared to the standard KFBI method. Being quite radical, the first method effectively reduces the KFBI method's iteration count to one, thereby considerably decreasing the running time, though at some cost to precision (relative to the KFBI method). However, as will be demonstrated in subsequent numerical experiments, this approach still achieves reasonably good numerical accuracy. The second method offers a balanced approach, acting as a preprocessing solution for setting the initial value in the KFBI iterative process. This method reduces a portion of the iteration count without compromising precision (relative to the KFBI method). Both methods are proven feasible and valuable through the upcoming numerical experiments. Additionally, due to the relatively simple architecture of the neural network, the time required for a single inference is negligible.
\item The operator learning approach based on the KFBI method, as proposed in this work, effectively reduces the problem's dimensionality. When addressing equation \eqref{one_GPU:PDE} for d=2, the input and output of the operator $\mathcal{S}_{\mathcal{L}, \Omega}$ (or $\mathcal{S}_{\mathcal{L}}$, $\mathcal{S}_{\Omega}$), approximated by the neural network $\mathcal{N}_{\mathcal{L}, \Omega}$ (or $\mathcal{N}_{\mathcal{L}}$, $\mathcal{N}_{\Omega}$), are functions on a one-dimensional manifold, and thus are naturally represented by vectors with $M$ components. This aspect of the method significantly aids in solving elliptic PDEs on $\Omega$ by learning mappings between functions defined on \(\partial \Omega\), highlighting a significant advantage of the approach.

\item The training of the neural network $\mathcal{N}_{\mathcal{L}, \Omega}$ enables the solution of a wide range of equations \eqref{one_GPU:PDE}, where both the non-homogeneous term $f$ and boundary conditions $g_D$ can vary. The applicability of neural networks $\mathcal{N}_{\mathcal{L}}$, $\mathcal{N}_{\Omega}$ and $\mathcal{N}$ is further extended, demonstrating their powerful range, which stands as another distinctive feature of this method. These models are pivotal in solving PDEs, as evidenced in Section \ref{Numerical Experiments}.

\item The proposed method can be synergistically combined with other techniques, such as multigrid solvers and GPU acceleration, to enhance solution efficiency.
\end{itemize}
\end{remark}

\section{Numerical Experiments} \label{Numerical Experiments}
This section presents numerical results for various classes of equations, including Laplace, Poisson, Stokes, modified Helmholtz, and Naiver equations. The focus is on the application of the hybrid KFBI method, detailing its operational mechanism and demonstrating its superiority in terms of numerical accuracy and iteration count. Since the neural network's structure, loss function, and training process have been elaborated in the preceding section, emphasis here is placed on post-training neural network applications. Excluding the training phase, programming in \textbf{C} and \textbf{C++} is utilized, with the libtorch library on the \textbf{C++} platform facilitating network inference computations. The examples in section \ref{Laplace Equations} are tested on a machine with a CPU `12th Gen Intel(R) Core(TM)i5-12600KF 3.70 GHZ' while the other examples are tested on a machine with a CPU `11th Gen Inter(R) Core(TM) i5-1135G7 @ 2.40GHz'.

For the KFBI method implementations, the Richardson iteration is employed as equation \eqref{one_GPU:richardson2} with $\gamma = 0.75$, setting the initial value to $\varphi_0 = 2g_D$ and the relative tolerance to 1E-8, in line with research \cite{ying2007kernel}. Unless otherwise specified, the parameters regarding the iteration stages mentioned above will be applied in each numerical experiment. Alphanumeric symbol definitions, such as $I$, $J$, and $M$, align with those in the second paragraph of section \ref{Operator Learning}. The numerical solution error is calculated by evaluating the norm of the difference between the numerical and exact solutions at all grid points in domain $\Omega$, with the error typically referring to absolute error unless stated otherwise.

\subsection{Hybrid KFBI without Parameter Component} \label{Hybrid-KFBI without Parameter Component}

\subsubsection{Laplace Equations} \label{Laplace Equations}
We focus on the post-training operations of the neural network model, excluding the training process, and this principle is consistently applied thereafter. In this context, the Laplace equation is considered with $\mathcal{L} = \Delta$ and the fixed homogeneous term $f = 0$. The boundary $\partial \Omega$ is defined by the set $\{(x, y): x = c_x + r_a \cos(\alpha) \cos(t) - r_b \sin(\alpha) \sin(t),\ y = c_y + r_a \sin(\alpha) \cos(t) - r_b \cos(\alpha) \sin(t)\ \mbox{for}\ t \in [0, 2\pi)\}$, where the ellipse, characterized as rotated, is defined by $c_x = 0.2, c_y = 0.4, r_a = 1.0, r_b = 0.5$, and $\alpha = \frac{\pi}{7}$. The bounding box $\mathcal{B}$ for the interface problem is established as $\mathcal{B} = [-1.2,1.2] \times [-1.2,1.2]$. The number of discrete points along the boundary curve $M$ is determined as $M = \max\{I, J\}$. To demonstrate the network's generalization capability, the exact solutions chosen for testing in the following sections were not included in the training data, a principle consistently upheld in subsequent examples.

We choose the boundary conditions given by different exact solutions to solve the Laplace equation. Table \ref{label1_1_1} displays the results obtained for Laplace Equation 1, where the exact solution \(u(x, y) = \exp(x) \cos(y) + \exp(y) \sin(x)\) is applied within the elliptic domain.
\begin{table}[ht]
    \centering
    \begin{tabular}{|c|c|c|c|c|} \hline 
         Grid size ($I\times J$) & $128 \times 128$ & $256 \times 256$ & $512 \times 512$ & $1024 \times 1024$ \\ \hline 
         $L_{\infty}$ error for Strategy 1 & 5.4E-3 & 2.2E-3 & 1.3E-3 & 1.7E-3\\ \hline
         $L_{\infty}$ error for standard KFBI & 4.0E-5 & 1.4E-5 & 1.7E-6 & 1.4E-7\\ \hline
         $L_{\infty}$ error for Strategy 2 & 4.0E-5 & 1.4E-5 & 1.7E-6 & 1.4E-7 \\ \hline
         Iterations (standard KFBI) &  26&  26&  26& 26\\ \hline 
         Iterations (Strategy 2)& 11&  12&  12& 11  \\ \hline
         
         Running time of 
         & \multirow{2}{*}{0.04154\bigstrut}
         & \multirow{2}{*}{0.1120\bigstrut}
         & \multirow{2}{*}{0.4010\bigstrut}
         & \multirow{2}{*}{1.513\bigstrut} \\
         standard KFBI (s) & & & & \\ \hline
         
         Running time of 
         & \multirow{2}{*}{0.02233\bigstrut}
         & \multirow{2}{*}{0.05463\bigstrut}
         & \multirow{2}{*}{0.1997\bigstrut}
         & \multirow{2}{*}{0.8390\bigstrut} \\
         Strategy 2 (s) & & & & \\ \hline
         
         Time saved (Strategy 2) & 46\%& 51\%& 50\%& 45\%\\ \hline
    \end{tabular}
    \caption{Result of Laplace equation 1: comparison of accuracy and efficiency in solving Laplace equation 1 using different methods.}
    \label{label1_1_1}
\end{table}

\begin{remark}
In table \ref{label1_1_1}, the second-row records results where the model's inference outputs are directly employed as the density function for solving the corresponding interface problem, aligning with Strategy 1 detailed in section \ref{Hybrid KFBI Method Based on Trained Models}. The last seven rows of the table compare the outcomes of the standard KFBI method with those achieved by initiating the KFBI iterative process using the model's inference results, a method consistent with Strategy 2 in the same section. It is noted that the precision of the numerical solutions obtained through this approach closely parallels that of the standard KFBI method, attributable to the identical relative tolerance set for the iterative processes. The subsequent formatting of the tables in section \ref{Hybrid-KFBI without Parameter Component} will remain consistent.
\end{remark}

Table \ref{label1_1_3} displays the results for Laplace Equation 2, characterized by the exact solution $u(x, y) = \sin(3x)\sinh(3y) + 0.5 \cosh(x) \cos(y)$, applied within the elliptic domain.
\begin{table}[ht]
    \centering
    \begin{tabular}{|c|c|c|c|c|} \hline 
         Grid size ($I\times J$) & $128 \times 128$ & $256 \times 256$ & $512 \times 512$ & $1024 \times 1024$ \\ \hline 
         $L_{\infty}$ error for Strategy 1 & 2.7E-3 & 1.4E-3 & 7.0E-4 & 8.6E-4\\ \hline
         $L_{\infty}$ error for standard KFBI & 2.0E-4 & 5.2E-5 & 1.3E-5 & 3.3E-6\\ \hline
         $L_{\infty}$ error for Strategy 2 & 2.0E-4 & 5.2E-5 & 1.3E-5 & 3.3E-6\\ \hline
         Iterations (standard KFBI) &  26&  26&  26& 26\\ \hline 
         Iterations (Strategy 2) &  11&  10&  12& 9\\ \hline 
         
         Running time of 
         & \multirow{2}{*}{0.04230\bigstrut}
         & \multirow{2}{*}{0.1144\bigstrut}
         & \multirow{2}{*}{0.4040\bigstrut}
         & \multirow{2}{*}{1.545\bigstrut} \\
         standard KFBI (s) & & & & \\ \hline
         
         Running time of 
         & \multirow{2}{*}{0.02296\bigstrut}
         & \multirow{2}{*}{0.05020\bigstrut}
         & \multirow{2}{*}{0.2023\bigstrut}
         & \multirow{2}{*}{0.7450\bigstrut} \\
         Strategy 2 (s) & & & & \\ \hline
        
         Time saved (Strategy 2) & 46\%& 56\%& 50\%& 52\%\\ \hline
    \end{tabular}
    \caption{Result of Laplace equation 2: comparison of accuracy and efficiency in solving the Laplace equation 3 using different methods.}
    \label{label1_1_3}
\end{table}

Results for Laplace Equation 3, featuring the exact solution $u(x, y) = \sin(2.5x) \sinh(2.5y)$ and applied to the elliptic domain, are detailed in Table \ref{label1_1_4}.
\begin{table}[ht]
    \centering
    \begin{tabular}{|c|c|c|c|c|} \hline 
         Grid size ($I\times J$) & $128 \times 128$ & $256 \times 256$ & $512 \times 512$ & $1024 \times 1024$ \\ \hline 
         $L_{\infty}$ error for Strategy 1 & 2.0E-3 & 9.6E-4 & 6.0E-4 & 1.0E-3 \\ \hline
         $L_{\infty}$ error for standard KFBI & 9.1E-5 & 3.7E-5 & 4.7E-6 & 1.2E-6 \\ \hline
         $L_{\infty}$ error for Strategy 2 & 9.1E-5 & 3.7E-5 & 4.7E-6 & 1.2E-6\\ \hline
         Iterations (standard KFBI) &  25&  25&  25& 25\\ \hline 
         Iterations (Strategy 2)&  11&  11&  14& 12\\ \hline 
         
         Running time of 
         & \multirow{2}{*}{0.04061\bigstrut}
         & \multirow{2}{*}{0.1099\bigstrut}
         & \multirow{2}{*}{0.3988\bigstrut}
         & \multirow{2}{*}{1.510\bigstrut} \\
         standard KFBI (s) & & & & \\ \hline
         
         Running time of 
         & \multirow{2}{*}{0.02237\bigstrut}
         & \multirow{2}{*}{0.05335\bigstrut}
         & \multirow{2}{*}{0.2289\bigstrut}
         & \multirow{2}{*}{0.8460\bigstrut} \\
         Strategy 2 (s) & & & & \\ \hline
         
         Time saved (Strategy 2) & 45\%& 51\%& 43\%& 44\%\\ \hline
    \end{tabular}
    \caption{Result of Laplace equation 3: comparison of accuracy and efficiency in solving Laplace equation 4 using different methods.}
    \label{label1_1_4}
\end{table}

The experimental data reveal that adopting Strategy 1 enables the accurate resolution of the corresponding Laplace equation with the model's assistance, resulting in a substantial reduction in the time required for solving PDEs, approximately $10\%$ of that required by the standard KFBI method. Moreover, implementing Strategy 2 significantly reduces iterative time (by about $50\%$) while maintaining precision comparable to the standard KFBI method.

\subsubsection{Two-Dimensional Stokes Equations}
In this section, 2D Stokes equation with Dirichlet boundary condition is defined as
\begin{equation} \label{eq4.1.3.1}
\begin{array}{rll}
-\Delta \mathbf{u}+\nabla p =\mathbf{f}, & & \text { in } \Omega^{+} \cup \Omega^{-}, \\
\nabla \cdot \mathbf{u}=0, & & \text { in } \Omega^{+} \cup \Omega^{-}, \\
\mathbf{u}=\mathbf{g_D}, & & \text { on } \partial \Omega,
\end{array}
\end{equation}
In the context presented, $\mathbf{u}=\left(u^{(1)}, u^{(2)}\right)^T$ and $p$ denote the unknown fluid velocity and pressure functions, respectively, while $\mathbf{f}=\left(f^{(1)}, f^{(2)}\right)^T$ represents the external force function.

The application of the KFBI method to solve the Stokes equation, as elaborated in the paper by Dong et al. \cite {dong2023second}, needs to be delved into here. This equation's solution $(\mathbf{u}, p)$ is expressible in terms of volume and boundary integrals, involving the density function $\varphi$, which solves the corresponding boundary integral equation. The right-hand side of this boundary integral equation is denoted as \(\widetilde{\mathbf{g_D}}\). Terms such as $\mathcal{S}_{\mathcal{L}, \Omega}$ (or $\mathcal{S}_{\mathcal{L}}$, $\mathcal{S}_{\Omega}$) and $\mathcal{N}_{\mathcal{L}, \Omega}$ (or $\mathcal{N}_{\mathcal{L}}$, $\mathcal{N}_{\Omega}$) retain their previous meanings. Notably, $\widetilde{\mathbf{g_D}}$ becomes a vector function \((\widetilde{g_D}^{(1)},\widetilde{g_D}^{(2)})\), and a similar transformation applies to \(\mathbf{\varphi}\). However, this does not introduce complications in the network's construction and training, as a tensor of size $(M, 2)$ can be conveniently reshaped into a vector of length $2M$. For this example, the absolute tolerance for iteration is set at 1E-8, and the initial value in the standard KFBI is $\mathbf{\varphi}_0 = (0, 0)^T$.

The boundary $\partial \Omega = \{(x, y): x = c_x + r_a cos(\alpha) cos(t) - r_b sin(\alpha) sin(t)\ \mbox{and}\ y = c_y + r_a sin(\alpha) cos(t) - r_b cos(\alpha) sin(t)\ \mbox{for}\ t \in [0, 2\pi) \}$ with $c_x = 0, c_y = 0, r_a = 1.0, r_b = 0.6$ and $\alpha = 0$ such that $\Omega$ is an ellipse. The bounding box $\mathcal{B}$ for the interface problem is set to be $\mathcal{B} = [-1.2,1.2] \times [-1.2,1.2]$. The discrete number of the boundary curve $M$ is set as $M = max\{I, J\}$.

Table \ref{label1_3} displays the results for Stokes Equation 1, using the exact solution specified in equation \eqref{eq4.1.3.2}, applied to the elliptic domain.

\begin{equation} \label{eq4.1.3.2}
\begin{cases} 
u^{(1)} = [(x^2 + y^2) \cos(7 \arctan(\frac{y}{x})) + 3.5 (1 - (x^2 + y^2)) \cos(5 \arctan(\frac{y}{x}))] (x^2 + y^2) ^ {\frac{5}{2}} \\
u^{(2)} = -3.5 (1 - (x^2 + y^2)) \sin(5\arctan(\frac{y}{x})) (x^2 + y^2) ^ {\frac{5}{2}}\\
p = -14 \cos(6 \arctan(\frac{y}{x}) ) (x^2 + y^2) ^ 3
\end{cases}
\end{equation}

\begin{table}[ht]
    \centering
    \begin{tabular}{|c|c|c|c|c|} \hline 
         Grid size $(I \times J)$ & $128 \times 128$ & $256 \times 256$ &$512 \times 512$   &$1024 \times 1024$\\ \hline 
         $L_{\infty}$ error for Strategy 1 & $\begin{bmatrix} 1.4E-3 \\ 1.1E-2 \\ 1.3E-2 \\ \end{bmatrix}$ 
         & $\begin{bmatrix} 1.1E-4 \\ 1.2E-4 \\ 1.3E-3 \\ \end{bmatrix}$
         & $\begin{bmatrix} 2.1E-4 \\ 2.3E-4 \\ 1.9E-3 \\ \end{bmatrix}$
         & $\begin{bmatrix} 1.9E-4 \\ 1.7E-4 \\ 3.6E-3 \\ \end{bmatrix}$
         \\ \hline
         $L_{\infty}$ error for KFBI & $\begin{bmatrix} 2.5E-4 \\ 5.2E-4 \\ 3.9E-3 \\ \end{bmatrix}$ 
         & $\begin{bmatrix} 6.3E-5 \\ 9.4E-5 \\ 6.0E-4 \\ \end{bmatrix}$ 
         & $\begin{bmatrix} 1.3E-5 \\ 2.4E-5 \\ 1.7E-4 \\ \end{bmatrix}$
         & $\begin{bmatrix} 3.1E-6 \\ 6.2E-6 \\ 4.5E-5 \\ \end{bmatrix}$ \\ \hline
         $L_{\infty}$ error for Strategy 2 & $\begin{bmatrix} 2.5E-4 \\ 5.2E-4 \\ 3.9E-3 \\ \end{bmatrix}$ 
         & $\begin{bmatrix} 6.3E-5 \\ 9.4E-5 \\ 6.9E-5 \\ \end{bmatrix}$
         & $\begin{bmatrix} 1.3E-5 \\ 2.4E-5 \\ 1.7E-4 \\ \end{bmatrix}$
         & $\begin{bmatrix} 3.1E-6 \\ 6.2E-6 \\ 4.5E-5 \\ \end{bmatrix}$
         \\ \hline
         Iterations (KFBI) &  20&  20&20 &       20\\ \hline 
         Iterations (Strategy 2)&  10&  10&12 &12\\ \hline 
         Running time of &  1.368&  5.404&23.84 &119.3\\ KFBI (s) & & & & \\ \hline 
         Running time of &  0.7433&  2.939&14.08 &68.40 \\ 
         Strategy 2 (s) & & & & \\ \hline
         Time saved (Strategy 2) & 46\%& 46\%& 41\%& 43\% \\ \hline
    \end{tabular}
    \caption{Result of 2D Stokes equation 1: comparison of accuracy and efficiency in solving 2D Stokes equation 1 using different methods. The error vector is computed by the numerical solution $(u_{\mbox{numerical}}^{(1)}, u_{\mbox{numerical}}^{(2)}, p_{\mbox{numerical}})$ and the exact solution $(u_{\mbox{exact}}^{(1)}, u_{\mbox{exact}}^{(2)}, p_{\mbox{exact}})$.}
    \label{label1_3}
\end{table}

From this table, it can be observed that our model significantly enhances the efficiency of solving the Stokes equation. When employing Strategy 1, we achieve high-precision numerical solutions. Moreover, timing experiments indicate that the time required to solve the equation on grids of size $128\times128$, $256\times256$, $512 \times 512$ and $1024 \times 1024$ using Strategy 1 is merely 0.1280s, 0.3862s, 1.650s and 6.103s, respectively. This leads to a substantial reduction in computation time, approximately 90\%! On the other hand, when employing Strategy 2, we can reduce the running time by almost half without compromising precision compared to the standard KFBI method.

\subsection{Hybrid KFBI on Parametric PDEs} \label{Hybrid-KFBI on Parametric PDEs}

\subsubsection{Poisson Equations on Star-Shaped Domains} \label{Poisson Equations on Star-Shaped Domains}
In this section, $\mathcal{L}$ is designated as $\Delta$, with $f$ being freely chosen as the inhomogeneous term of the Poisson equation. The boundary defining the star-shaped domain $\Omega$ is characterized by $\partial \Omega = \{(x, y): x = (1.0 - S_c + S_c \cos(S_m t)) \cos(t),\ y = (1.0 - S_c + S_c \cos(S_m t)) \sin(t)\ \mbox{for}\ t \in [0, 2\pi) \}$, where $S_m \in \{3, 4, 5, 6\}$ and $S_c \in [0.05, 0.20]$ are parameters included in the input part of networks. The bounding box $\mathcal{B}$ for the interface problem is established as $\mathcal{B} = [-1.5,1.5] \times [-1.5,1.5]$. The discrete number of points along the boundary curve $M$ is determined as $M = \max\{I, J\}$, with $I = J = 256$ in this example.

Table \ref{label2_4} presents the results for Poisson Equation 3 with the exact solution $u(x, y) = \exp(x) \\ \cos(y) + \exp(y) \sin(x) + \exp(0.6 x + 0.8 y)$, applied to star-shaped domains varying in parameters $S_m, S_c$.

\begin{table}[ht]
    \centering
    \begin{tabular}{|c|c|c|c|c|} \hline 
         $(S_m, S_c) $ & $(3, 0.20)$& $(4, 0.10) $& $(5, 0.05)$& $(6, 0.15) $\\ \hline 
         $L_{\infty}$ error for Strategy 1 & 4.2E-3 & 5.8E-3 & 3.3E-3 & 7.8E-3 \\ \hline
         $L_{\infty}$ error for standard KFBI & 1.2E-5 & 1.3E-5 & 1.3E-5 & 5.0E-5 \\ \hline
         $L_{\infty}$ error for Strategy 2 & 1.2E-5 & 1.3E-5 & 1.3E-5 & 5.1E-5 \\ \hline
         Iterations (standard KFBI) &  26&  26&25& 27\\ \hline 
         Iterations (Strategy 2) & 14&  13&14&17\\  \hline 
         
         Network inference cost
         & \multirow{2}{*}{0.69\bigstrut}
         & \multirow{2}{*}{0.67\bigstrut}
         & \multirow{2}{*}{0.57\bigstrut}
         & \multirow{2}{*}{0.63\bigstrut} \\
         (converted to iteration counts) & & & & \\ \hline
          
         Saved iterations & 43\%& 47\%& 42\%&35\%\\ \hline
    \end{tabular}
    \caption{Result of Poisson equation 1: comparison of accuracy and efficiency in solving Poisson equation 1 using different methods.}
    \label{label2_4}
\end{table}

\begin{remark}
In Table \ref{label2_4}, the second-row records results where the model's inference outputs are directly used as the density function for solving the corresponding interface problem, aligning with Strategy 1 in section \ref{Hybrid KFBI Method Based on Trained Models}. The last six rows of the table compare outcomes between the standard KFBI method and those achieved by initiating the KFBI iterative process using the model's inference results, as outlined in Strategy 2 of the same section. Notably, the penultimate row in the table represents the conversion of network inference time into an equivalent number of iterations, offering an alternative perspective on observing the saving of iteration time. The subsequent formatting of the tables in sections \ref{Hybrid-KFBI on Parametric PDEs} will remain consistent.
\end{remark}

The table illustrates that, despite including parameter components, the trained model retains its accuracy and robust generalization capabilities. The error associated with Strategy 1 is consistently maintained at magnitudes of $10^{-3}$, and its execution time is notably reduced compared to the standard KFBI method, equivalent to a saving of about 22.6 iterations (considering that Strategy 1 entails one network inference and a single call to the interface problem solver). In the case of Strategy 2, the model demonstrates the ability to reduce approximately 45\% of the iteration steps required by the standard KFBI method without compromising precision.

\subsubsection{Stokes Equations on Ellipses}
In this section, Stokes Equation \eqref{eq4.1.3.1} is considered. The boundary defining the elliptic domain $\Omega$ is described by $\partial \Omega = \{(x, y): x = r_a \cos(t),\ y = r_b \sin(t)\ \mbox{for}\ t \in [0, 2\pi) \}$, with parameters $r_a, r_b \in [0.8, 1.2]$. The coordinates on parameterized $\partial \Omega$ appear in the networks. The bounding box $\mathcal{B}$ for the interface problem is established as $\mathcal{B} = [-1.5,1.5] \times [-1.5,1.5]$. The discrete number of points along the boundary curve $M$ is set at $M = \max\{I, J\}$, with $I = J = 128$ in this example, and the absolute tolerance for iteration is fixed at 1E-8.

Table \ref{label2_5} displays the results for Stokes Equation 2, using the exact solution given by equation \eqref{eq4.2.3.1}, and applied to ellipses with varying parameters $r_a, r_b$.
\begin{equation} \label{eq4.2.3.1}
\begin{cases} 
u^{(1)} = 2xy \\
u^{(2)} = 1 - (x^2 + y^2)\\
p = -4 y
\end{cases}
\end{equation}

\begin{table}[ht]
    \centering
    \begin{tabular}{|c|c|c|c|c|} \hline 
         $(r_a, r_b) $ & $(0.8, 0.8)$& $(1.0, 1.1) $& $(1.1, 1.0)$ & $(1.1, 1.1) $\\ \hline 
         $L_{\infty}$ error for Strategy 1 
         & $\begin{bmatrix} 7.6E-4 \\ 3.5E-4 \\ 9.7E-3 \\ \end{bmatrix}$ 
         & $\begin{bmatrix} 5.4E-4 \\ 7.7E-4 \\ 5.1E-3 \\ \end{bmatrix}$  
         &  $\begin{bmatrix} 5.0E-4 \\ 4.8E-4 \\ 5.5E-3 \\ \end{bmatrix}$ 
         &  $\begin{bmatrix} 7.6E-4 \\ 7.2E-4 \\ 6.2E-3 \\ \end{bmatrix}$ \\ \hline
         $L_{\infty}$ error for KFBI 
         &  $\begin{bmatrix} 5.2E-5 \\ 3.5E-5 \\ 1.3E-3 \\ \end{bmatrix}$
         &  $\begin{bmatrix} 3.5E-5 \\ 2.8E-5 \\ 6.9E-4 \\ \end{bmatrix}$ 
         &  $\begin{bmatrix} 2.5E-5 \\ 2.4E-5 \\ 5.5E-4 \\ \end{bmatrix}$ 
         &  $\begin{bmatrix} 2.9E-5 \\ 3.2E-5 \\ 5.5E-4 \\ \end{bmatrix}$ \\ \hline
         $L_{\infty}$ error for Strategy 2 
         &  $\begin{bmatrix} 5.2E-5 \\ 3.5E-5 \\ 1.3E-3 \\ \end{bmatrix}$ 
         &  $\begin{bmatrix} 3.5E-5 \\ 2.8E-5 \\ 6.9E-4 \\ \end{bmatrix}$ 
         &  $\begin{bmatrix} 2.5E-5 \\ 2.4E-5 \\ 5.5E-4 \\ \end{bmatrix}$ 
         &  $\begin{bmatrix} 2.9E-5 \\ 3.2E-5 \\ 5.5E-4 \\ \end{bmatrix}$ \\ \hline
         Iterations (KFBI) &  26&  26&26&       26\\ \hline 
         Iterations (Strategy 2) & 17&  14&15&15\\ \hline 
         
         Network inference cost
         & \multirow{2}{*}{0.19\bigstrut}
         & \multirow{2}{*}{0.22\bigstrut}
         & \multirow{2}{*}{0.21\bigstrut}
         & \multirow{2}{*}{0.20\bigstrut} \\
         (converted to iteration counts) & & & & \\ \hline
         
         Save iterations & 34\%& 45\%& 41\%& 42\%\\ \hline
    \end{tabular}
    \caption{Result of 2D Stokes equation 2: comparison of accuracy and efficiency in solving 2D Stokes Equation 2 using different methods. The error vector is computed by the numerical solution $(u_{\mbox{numerical}}^{(1)}, u_{\mbox{numerical}}^{(2)}, p_{\mbox{numerical}})$ and the exact solution $(u_{\mbox{exact}}^{(1)}, u_{\mbox{exact}}^{(2)}, p_{\mbox{exact}})$.}
    \label{label2_5}
\end{table}
From the data presented in this table, a conclusion akin to that drawn in section \ref{Poisson Equations on Star-Shaped Domains} is derived regarding our trained model. It is observed that employing the model's direct inference results as the exact solution for the density function (namely, Strategy 1) yields sufficiently high accuracy.

\subsubsection{Modified Helmholtz Equations} \label{Modified Helmholtz Equations}
\begin{figure}[htb]
\centering
\begin{minipage}[t]{0.48\textwidth}
\centering
\includegraphics[width=0.95\textwidth]{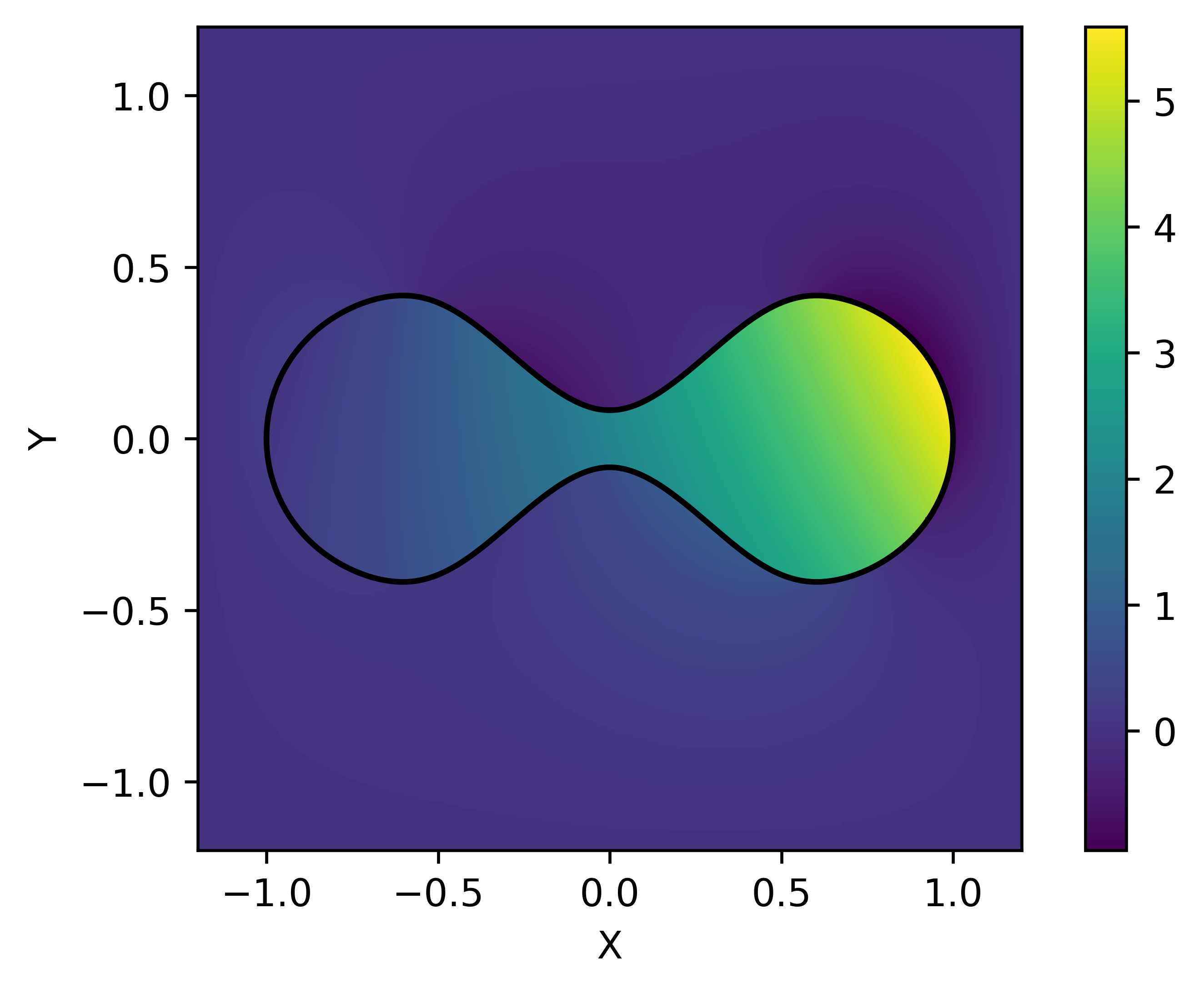}
\captionof*{figure}{Numerical solution by Strategy 1.}
\end{minipage}
\begin{minipage}[t]{0.48\textwidth}
\centering
\includegraphics[width=0.95\textwidth]{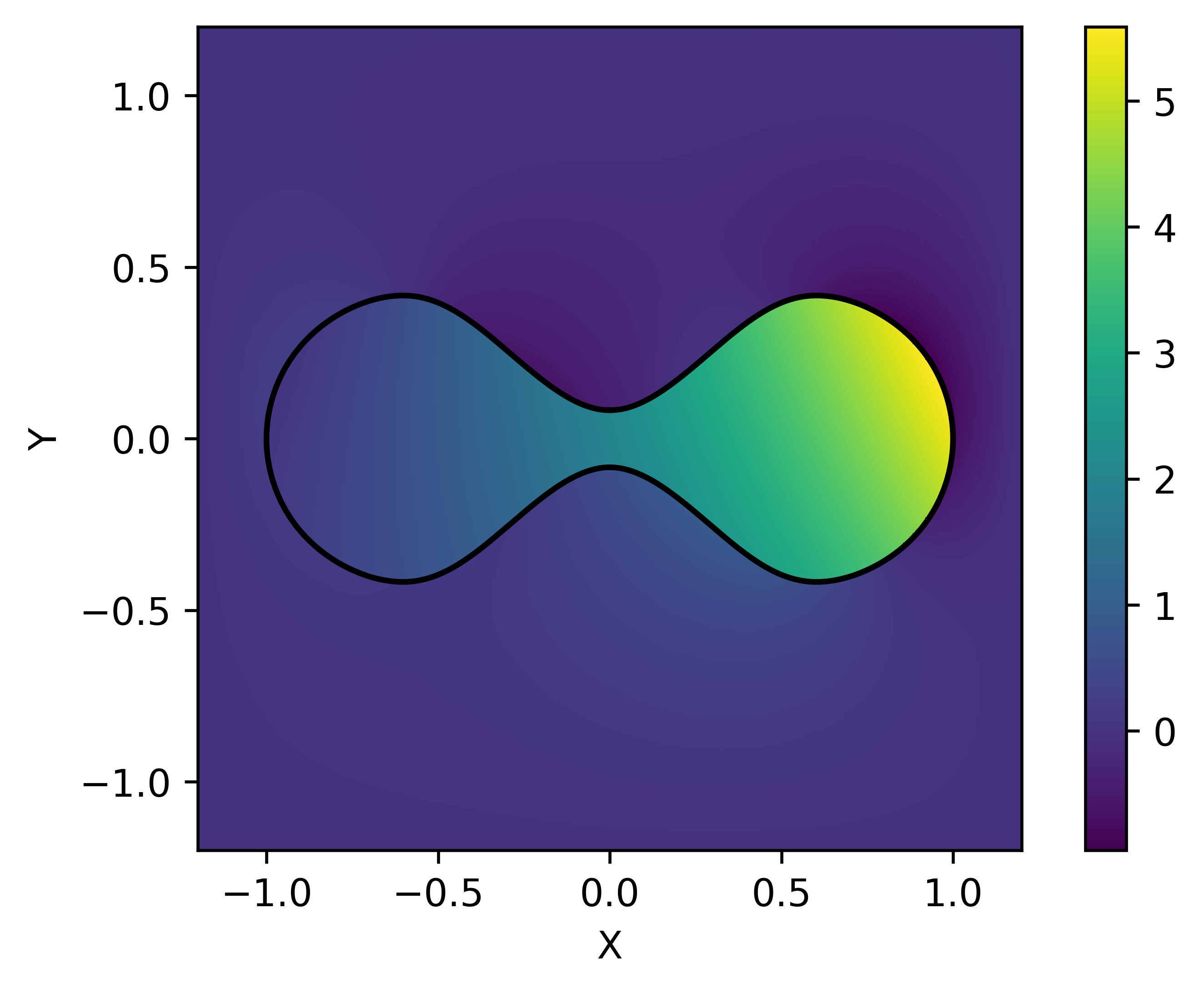}
\captionof*{figure}{Numerical solution by Strategy 2.}
\end{minipage}
\caption{Numerical solutions of Modified Helmholtz equation 1 given by Strategy 1 and Strategy 2 in grid $256 \times 256$. Note that the figures show the results given by solving the corresponding interface problem in KFBI whose interior solution is the desired results for the original PDE.}
\label{fig:Modified Helmholtz}
\end{figure}

In this section, the operator $\mathcal{L}$ is defined as $\Delta - \kappa \mathcal{I}$, where $\mathcal{I}$ represents the identity operator, and $\kappa > 0$ serves as the coefficient for the modified Helmholtz equation. The domain $\Omega$ is designated a dumbbell-shaped area, as depicted in figure \ref{fig:Modified Helmholtz}. Specifically, $\partial \Omega$ is a periodic $C^2$ cubic spline curve, passing through control points detailed in Appendix \ref{appen:Control Points for Dumbbel-Shaped Domain}. The coefficient $\kappa$, ranging between 1 and 5, is integrated into the input part of the networks. The bounding box $\mathcal{B}$ for the interface problem is established as $\mathcal{B} = [-1.2,1.2] \times [-1.2,1.2]$. The number of discrete points along the boundary curve $M$ is set at $M = 92$, with $I = J = 256$ in this example. The absolute tolerance for iteration is fixed at 1E-8, and the initial value for the standard KFBI method is set as $\mathbf{\varphi}_0 = (0, 0)^T$.

Table \ref{label3_1_1} presents the results for modified Helmholtz Equation 1, featuring the exact solution \(u(x, y) = \exp(x) \cos(y) + \exp(y) \sin(x) + \exp(0.6x + 0.8y)\), applied to the dumbbell-shaped domain with varying values of $\kappa$. For illustration, numerical solutions obtained through Strategy 1 and Strategy 2, using a grid size of $256 \times 256$ and a coefficient $\kappa = 3$, are depicted in figure \ref{fig:Modified Helmholtz}.
\begin{table}[ht]
    \centering
    \begin{tabular}{|c|c|c|c|c|c|c|} \hline 
         $\kappa $ & 1.1 & 1.3& $\frac{\pi}{2}$ & 2 & 3 \\ \hline 
         $L_{\infty}$ error for Strategy 1 & 1.3E-3 & 1.4E-3 & 1.3E-3 & 1.2E-3 & 1.3E-3  \\ \hline
         $L_{\infty}$ error for standard KFBI & 1.5E-4 & 1.8E-4 & 2.2E-4 & 2.8E-4 & 4.2E-4  \\ \hline
         $L_{\infty}$ error for Strategy 2 & 1.5E-4 & 1.8E-4 & 2.2E-4 & 2.8E-4 & 4.2E-4 \\ \hline
         Iterations (standard KFBI) & 57 & 56 & 55 & 53 & 49 \\ \hline 
         Iterations (Strategy 2) & 25 & 25 & 27 & 26 & 25 \\ \hline 
         
         Network inference cost
         & \multirow{2}{*}{0.18\bigstrut}
         & \multirow{2}{*}{0.19\bigstrut}
         & \multirow{2}{*}{0.16\bigstrut}
         & \multirow{2}{*}{0.20\bigstrut}
         & \multirow{2}{*}{0.19\bigstrut} \\
         (converted to iteration counts) & & & & & \\ \hline
         
         Save iterations & 56\% & 55\% & 51\% & 51\% & 49\% \\ \hline
    \end{tabular}
    \caption{Result of modified Helmholtz equation 1: comparison of accuracy and efficiency in solving modified Helmholtz equation 1 using different methods.}
    \label{label3_1_1}
\end{table}

The table demonstrates that the model trained in this example exhibits exceptional and consistent performance. For varying $\kappa$ values, Strategy 1 enables the achievement of satisfactory precision. Furthermore, the application of Strategy 2 significantly reduces the iteration time of the KFBI method. Notably, due to the domain's shape in this example, the iteration count for the standard KFBI method is considerably higher than in previous cases. However, the models outperform through both Strategy 1 and Strategy 2. Particularly with Strategy 1, corresponding to only about 1.2 iterations (involving one network inference and a single call to the interface problem solver), the method saves over $97\%$ of iteration time compared to the standard KFBI method, which requires over 50 iterations.

\subsubsection{Modified Helmholtz Equations on Parametric Domains} \label{Modified Helmholtz Equations on Parametric Domains}
In this section, we extend the framework presented in \ref{Modified Helmholtz Equations} by incorporating diversity in the computational domain. Specifically, beyond varying the parameter $\kappa$ within the range $[0.5, 3.0]$, the computational domain $\Omega$ can also undergo transformations involving rotation by an angle $\alpha$ and scaling by a factor $r$ relative to the original domain $\Omega_0$. Here, the boundary of domain $\Omega_0$ is similarly obtained through cubic spline interpolation of a set of control points (which is recorded in Appendix \ref{appen:Control Points for Dumbbel-Shaped Domain} for details), and $\alpha \in [0, 2 \pi]$, $r \in [0.6, 1]$. In this case, both $\kappa$ and the coordinates of $\partial \Omega$ along with the values of $\widetilde{g_D}$ at $M$ 
discrete points are the input of the network model. The number of discrete points along the boundary curve $M$ is set at $M = 46$, with $I = J = 128$ in this example. The absolute tolerance for iteration is fixed at 1E-6, and the initial value for the standard KFBI method is set as $\mathbf{\varphi}_0 = (0, 0)^T$. Other details correspond to the experiments in \ref{Modified Helmholtz Equations}.

\begin{figure}[htb]
\centering
\begin{minipage}[t]{0.48\textwidth}
\centering
\includegraphics[width=0.95\textwidth]{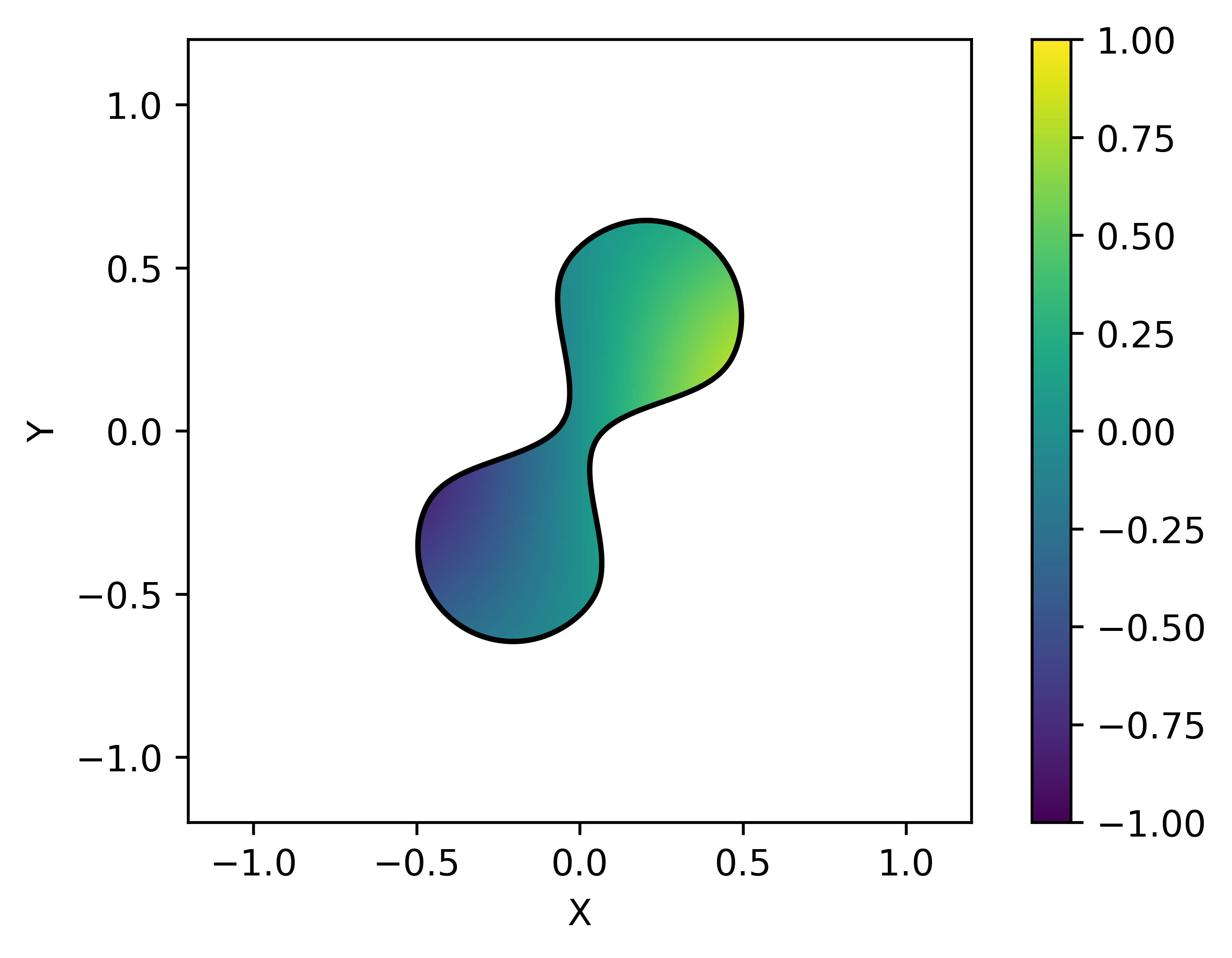}
\captionof*{figure}{Exact solution.}
\end{minipage}
\begin{minipage}[t]{0.48\textwidth}
\centering
\includegraphics[width=0.95\textwidth]{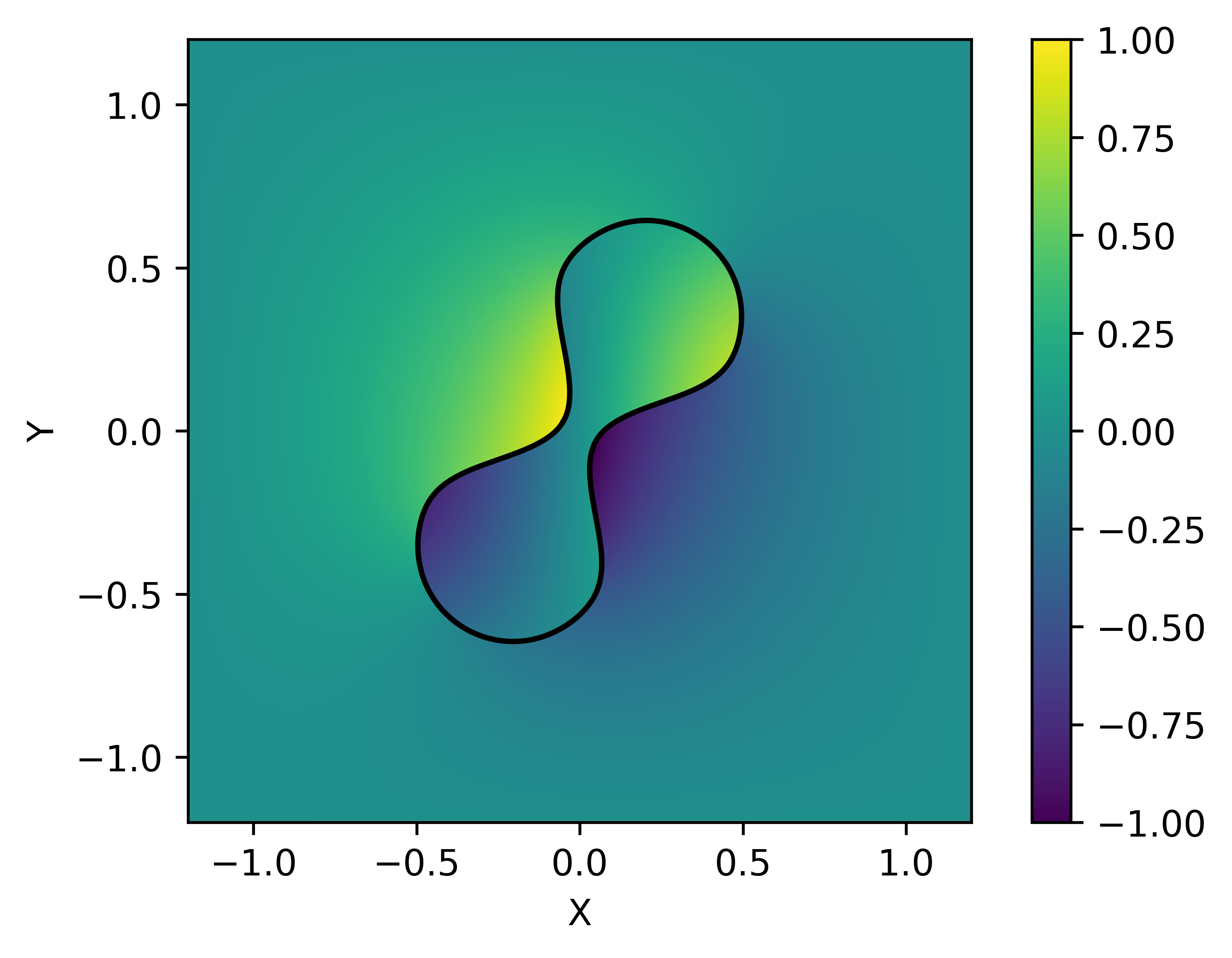}
\captionof*{figure}{Numerical solution by Strategy 1.}
\end{minipage}
\caption{Exact solution and Numerical solution of Modified Helmholtz equation 2 given by Strategy 1 in grid $128 \times 128$. Note that the right figure show the result given by solving the corresponding interface problem in KFBI whose interior solution is the desired result for the original PDE.}
\label{fig:Modified Helmholtz2}
\end{figure}

Table \ref{label_Cont_1} presents the results for modified Helmholtz Equation 2, featuring the exact solution \(u(x, y) = \sin(2.1 x) \cos(1.9 y) \), applied to the transformed dumbbell-shaped domain with varying values $\alpha$, $r$ and $\kappa$. For illustration, numerical solutions obtained through Strategy 1 and Strategy 2, using a grid size of $128 \times 128$ and coefficients $\alpha = \frac{\pi}{3}$, $r = 0.7$ and $\kappa = 2.8$, are depicted in figure \ref{fig:Modified Helmholtz2}.

\begin{table}[ht]
    \centering
    \begin{tabular}{|c|c|c|c|c|c|} \hline 
         $(\alpha, r, \kappa) $ & $(\frac{\pi}{3}, 0.7, 2.8)$ & $(\frac{\pi}{4}, 0.7, 2.8)$ & $(\frac{\pi}{6}, 0.71, 0.9)$ & $(\frac{\pi}{5}, 0.68, 1.3)$ \\ \hline 
         $L_2$ error for Strategy 1 & 1.1E-3 & 1.5E-3& 4.2E-3 & 1.6E-3 \\ \hline
         $L_{2}$ error for standard KFBI & 4.4E-5 & 4.8E-5 & 1.7E-5 & 2.2E-5 \\ \hline
         $L_{2}$ error for Strategy 2 & 4.4E-5 & 4.8E-5 & 1.7E-5 & 2.2E-5 \\ \hline
         Iterations (standard KFBI) & 50 & 49 & 54 & 54 \\ \hline 
         Iterations (Strategy 2) & 28 & 26 & 32 & 32  \\ \hline 
         
         Network inference cost
         & \multirow{2}{*}{0.86\bigstrut}
         & \multirow{2}{*}{0.81\bigstrut}
         & \multirow{2}{*}{0.83\bigstrut}
         & \multirow{2}{*}{0.79\bigstrut}
         \\
         (converted to iteration counts) & & & &\\ \hline
         
         Save iterations & 42\% & 45\% & 39\% & 39\% \\ \hline
    \end{tabular}
    \caption{Result of modified Helmholtz equation 2: comparison of accuracy and efficiency in solving modified Helmholtz equation 2 using different methods.}
    \label{label_Cont_1}
\end{table}

From the above results, it is evident that our model can effectively handle parametric PDEs, where not only the parameters of the equation can vary but also systematic variations can be applied to the computational domain. Figure \ref{fig:Modified Helmholtz2} above further demonstrates that our Strategy 1 exhibits remarkable accuracy, while its computational speed is astonishing, consuming less than $3\%$ of the time required by the traditional KFBI method.

\subsubsection{Naiver Equations} \label{Naiver Equations}
In this section, Naiver equation with Dirichlet boundary condition is defined as
\begin{equation} \label{eq4.3.2.1}
\begin{aligned}
\nabla \cdot \boldsymbol{\sigma}(\mathbf{u})+\mathbf{f} & =\mathbf{0}, & & \text { in } \Omega, \\
\mathbf{u} & =\mathbf{g_D}, & & \text {on} \partial \Omega,
\end{aligned}
\end{equation}
here $\mathbf{u}=\left(u^{(1)}, u^{(2)}\right)^T,\ \mathbf{f}=\left(f^{(1)}, f^{(2)}\right)^T$ represent displacement variable and external force, respectively. Stress tensor is
\begin{equation} \label{eq4.3.2.2}
\boldsymbol{\sigma}(\mathbf{u})=\lambda \nabla \cdot \mathbf{u} \mathbf{I}+2 \mu \boldsymbol{\epsilon}(\mathbf{u}),
\end{equation}
here $\boldsymbol{\epsilon}(\mathbf{u})=\frac{1}{2}\left(\nabla \mathbf{u}+(\nabla \mathbf{u})^T\right)$ is the linear strain tensor, $\mathbf{I}$ is the $2 \times 2$ identity matrix. La$\acute{\mbox{m}}$e coefficients $\lambda$ and $\mu$ is given by Young's modulus $E$ and Poisson's ratio $\nu$:
\begin{equation*}
\lambda=\frac{E \nu}{(1+\nu)(1-2 \nu)}, \quad \mu=\frac{E}{2(1+\nu)}.
\end{equation*}
By inserting equation \eqref{eq4.3.2.2} into equation \eqref{eq4.3.2.1}, we can obtain this PDE with unknown $\mathbf{u}$
\begin{equation} \label{eq4.3.2.3}
-\mu \Delta \mathbf{u}-(\lambda+\mu) \nabla(\nabla \cdot \mathbf{u})=\mathbf{f}, \quad \text { in } \Omega .
\end{equation}
The solution of the Naiver equation using the KFBI method, as detailed in the paper by Zhao et al.\cite{zhao2023kernel}, needs not to be elaborated upon here; readers interested in its implementation may consult the referenced paper. It is important to note that the solution $\mathbf{u}$ to this equation can be expressed through volume and boundary integrals, with the density function $\varphi$ resolving the corresponding boundary integral equation. In this work, the right-hand side of this boundary integral equation is continued to be denoted as $\widetilde{\mathbf{g_D}}$, derived from both $\mathbf{g_D}$ and $\mathbf{f}$. Terms such as $\mathcal{S}_{\mathcal{L}, \Omega}$ and $\mathcal{N}_{\mathcal{L}, \Omega}$ retain their previously established meanings.

\begin{figure}[htb]
\centering
\begin{minipage}[t]{0.45\textwidth}
\centering
\includegraphics[width=0.9\textwidth]{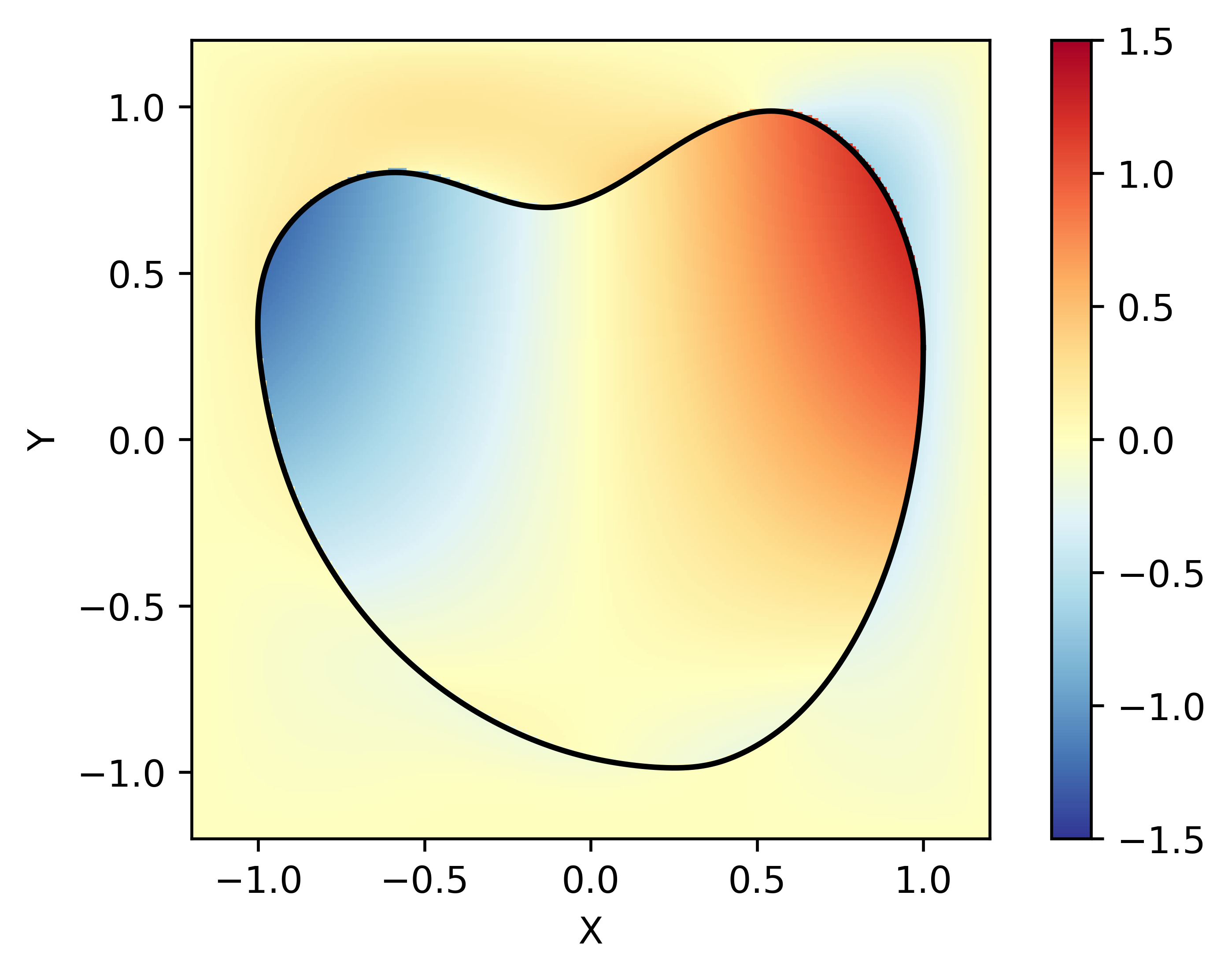}
\captionof*{figure}{Displacement $u^{(1)}$.}
\end{minipage}
\begin{minipage}[t]{0.45\textwidth}
\centering
\includegraphics[width=0.9\textwidth]{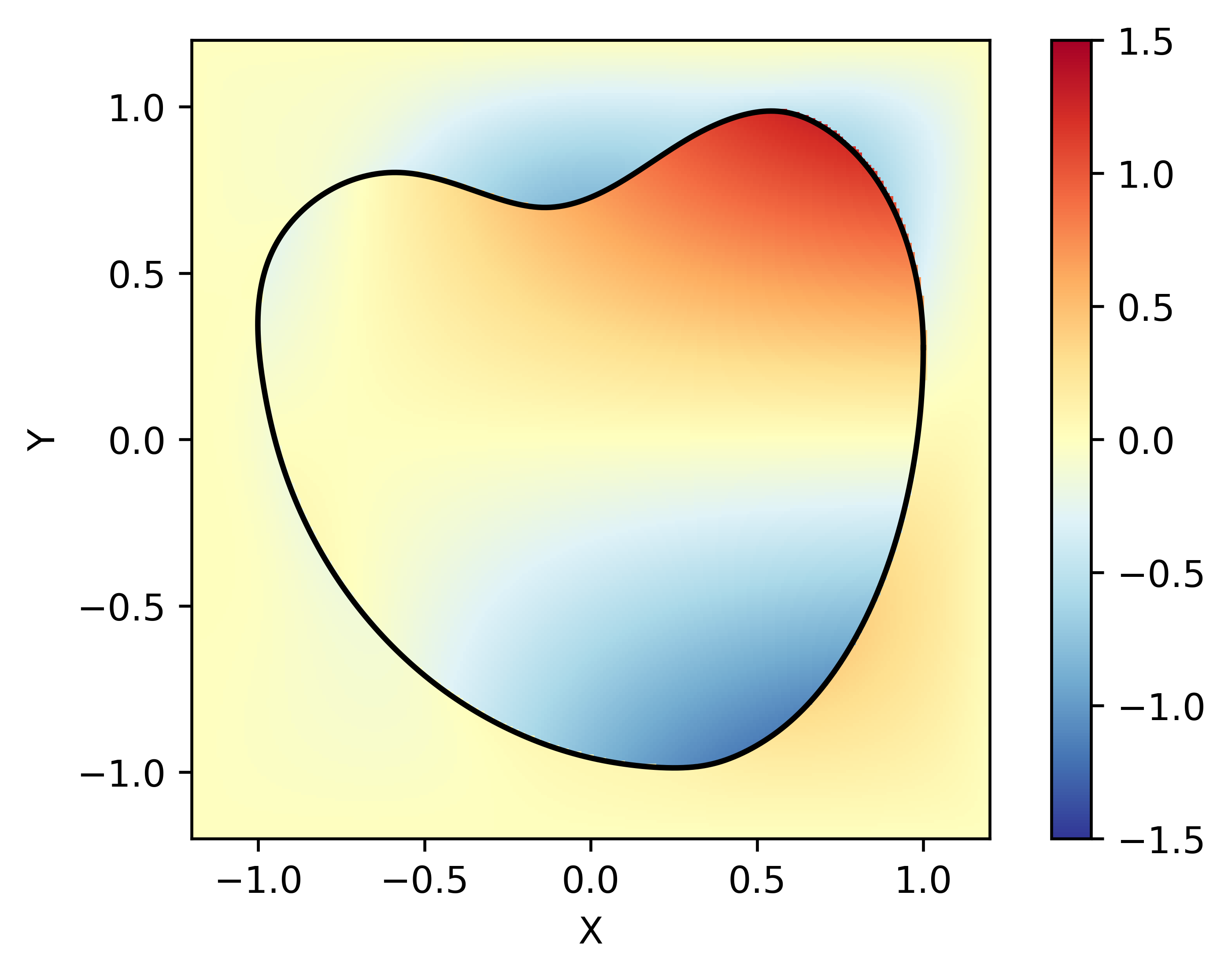}
\captionof*{figure}{Displacement $u^{(2)}$.}
\end{minipage}
\caption{Numerical solution of the 2D Naiver equation 1 given by Strategy 2 in grid $256 \times 256$. Note that the figures show the result given by solving the corresponding interface problem in KFBI whose interior solution is the desired result for the original PDE.}
\label{fig:naiver2D}
\end{figure}

This section defines the domain $\Omega$ as a heart-shaped area, as illustrated in figure \ref{fig:naiver2D}. $\partial \Omega$ is a periodic $C^2$ cubic spline curve, traversing control points detailed in Appendix \ref{appen:Control Points for Heart-Shaped Domain}. The parameters $E \in [10^8, 10^9]$ and $\nu \in [0.35, 0.45]$ are integrated into the input part of networks, covering the typical range of Young's modulus and Poisson's ratio for most plastics. The bounding box $\mathcal{B}$ for the interface problem is established as $\mathcal{B} = [-1.2,1.2] \times [-1.2,1.2]$. The number of discrete points on the boundary curve $M$ is set at $M = \max\{I, J\}$ with $I = J = 256$. Absolute tolerance for iteration is determined to be 1E-8, and the initial value for the standard KFBI method is set as $\phi_0 = (0, 0)^T$.

\begin{remark}
\begin{itemize}
\item In practice, $\log_{10} E$ is used in the networks instead of $E$.
\end{itemize}
\end{remark}

Table \ref{label3_1_2} presents the results for Naiver Equation 1, featuring exact solution $u^{(1)} = \sin x \cos y \\ + xy$ and $u^{(2)} = \cos x \sin y + xy$, applied to the heart-shaped domain with various $(E, \nu)$ parameter values. As an example, the numerical solution obtained through Strategy 2, employing a grid size of $256 \times 256$ and coefficients $(E, \nu) = (5.5\times 10^8, 0.4)$, is depicted in figure \ref{fig:naiver2D}.
\begin{table}[ht]
    \centering
    \begin{tabular}{|c|c|c|c|c|} \hline 
         $(E, \nu) $ & (3E08, 0.4) & (5.2E08, 0.35) & (5.5E08, 0.4) & (9E08, 0.36) \\ \hline 
         $L_{\infty}$ error for Strategy 1 & $\begin{bmatrix}  9.5E-4\\ 8.5E-4\\ \end{bmatrix}$  
         & $\begin{bmatrix}  5.1E-4\\ 4.9E-4\\ \end{bmatrix}$ 
         & $\begin{bmatrix}  9.6E-4\\ 8.5E-4\\ \end{bmatrix}$ 
         & $\begin{bmatrix}  1.8E-3\\ 1.1E-3\\ \end{bmatrix}$  \\ \hline
         $L_{\infty}$ error for KFBI & $\begin{bmatrix}  1.4E-4\\ 1.2E-4\\ \end{bmatrix}$ 
         & $\begin{bmatrix}  1.0E-4\\ 8.9E-5\\ \end{bmatrix}$ 
         & $\begin{bmatrix}  1.4E-4\\ 1.2E-4\\ \end{bmatrix}$ 
         & $\begin{bmatrix}  1.1E-4\\ 9.3E-5\\ \end{bmatrix}$  \\ \hline
         $L_{\infty}$ error for Strategy 2 & $\begin{bmatrix}  1.4E-4\\ 1.2E-4\\ \end{bmatrix}$ 
         & $\begin{bmatrix}  1.0E-4\\ 8.9E-5\\ \end{bmatrix}$ 
         & $\begin{bmatrix}  1.4E-4\\ 1.2E-4\\ \end{bmatrix}$ 
         & $\begin{bmatrix}  1.1E-4\\ 9.3E-5\\ \end{bmatrix}$ \\ \hline
         Iterations (KFBI) & 26 & 26 & 26 & 26\\ \hline 
         Iterations (Strategy 2) & 15 & 14 & 15 &  17 \\ \hline 
         
         Network inference time
         & \multirow{2}{*}{0.079\bigstrut}
         & \multirow{2}{*}{0.072\bigstrut}
         & \multirow{2}{*}{0.081\bigstrut}
         & \multirow{2}{*}{0.071\bigstrut} \\
         in iteration steps & & & & \\ \hline
         
         Save iterations & 42\% & 46\% & 42\% & 34\% \\ \hline
    \end{tabular}
    \caption{Result of Naiver equation 1: comparison of accuracy and efficiency in solving Naiver equation 1 using different methods. The error vector is computed by the numerical solution $(u_{\mbox{numerical}}^{(1)}, u_{\mbox{numerical}}^{(2)})$ and the exact solution $(u_{\mbox{exact}}^{(1)}, u_{\mbox{exact}}^{(2)})$.}
    \label{label3_1_2}
\end{table}

From the data in this table, a conclusion is derived from section \ref{Modified Helmholtz Equations} regarding our trained model. It becomes evident that applying direct model inference outcomes as the solution for the density function (Strategy 1) yields a sufficiently high level of accuracy.

\subsubsection{Naiver Equations on Domains with Boundaries Defined by Perturbed Control Points} \label{Naiver Equations on Domains with Boundaries Defined by Perturbed Control Points}
In this section, we continue to focus on the Naiver equations presented in \ref{Naiver Equations}. We allow for perturbations to be applied to certain control points, and incorporate the coordinates of these control points into the input of the model (note that $\partial \Omega$ is obtained through cubic spline interpolation of the control points). Specifically in this experience, the original domain is a heart-shaped domain whose boundary is obtained through cubic spline interpolation of 28 control points, identical to the boundary described in \ref{Naiver Equations}. We allow for perturbations in the x-direction of the $1st$, $14th$, and $27th$ control points, with distances of $\epsilon_1, \epsilon_{14}, \epsilon_{27}$, respectively. Here $\epsilon_i \in [-0.15, 0.15], i \in \{1, 14, 27\}$. As previously mentioned, the coordinates of these three points will be included in the model's input along with the values of $\widetilde{\mathbf{g_D}}$ at $M$ 
discrete points. The equation parameters $E$ and $\nu$ are fixed at 1E9 and 0.45, respectively. The bounding box $\mathcal{B}$ for the interface problem is established as $\mathcal{B} = [-1.5,1.5] \times [-1.5,1.5]$. The number of discrete points on the boundary curve $M$ is set at $M = \max\{I, J\}$ with $I = J = 128$. Absolute tolerance for iteration is determined to be 1E-8, and the initial value for the standard KFBI method is set as $\phi_0 = (0, 0)^T$.

\begin{figure}[htb]
\centering
\begin{minipage}[t]{0.22\textwidth}
\centering
\includegraphics[width=0.9\textwidth]{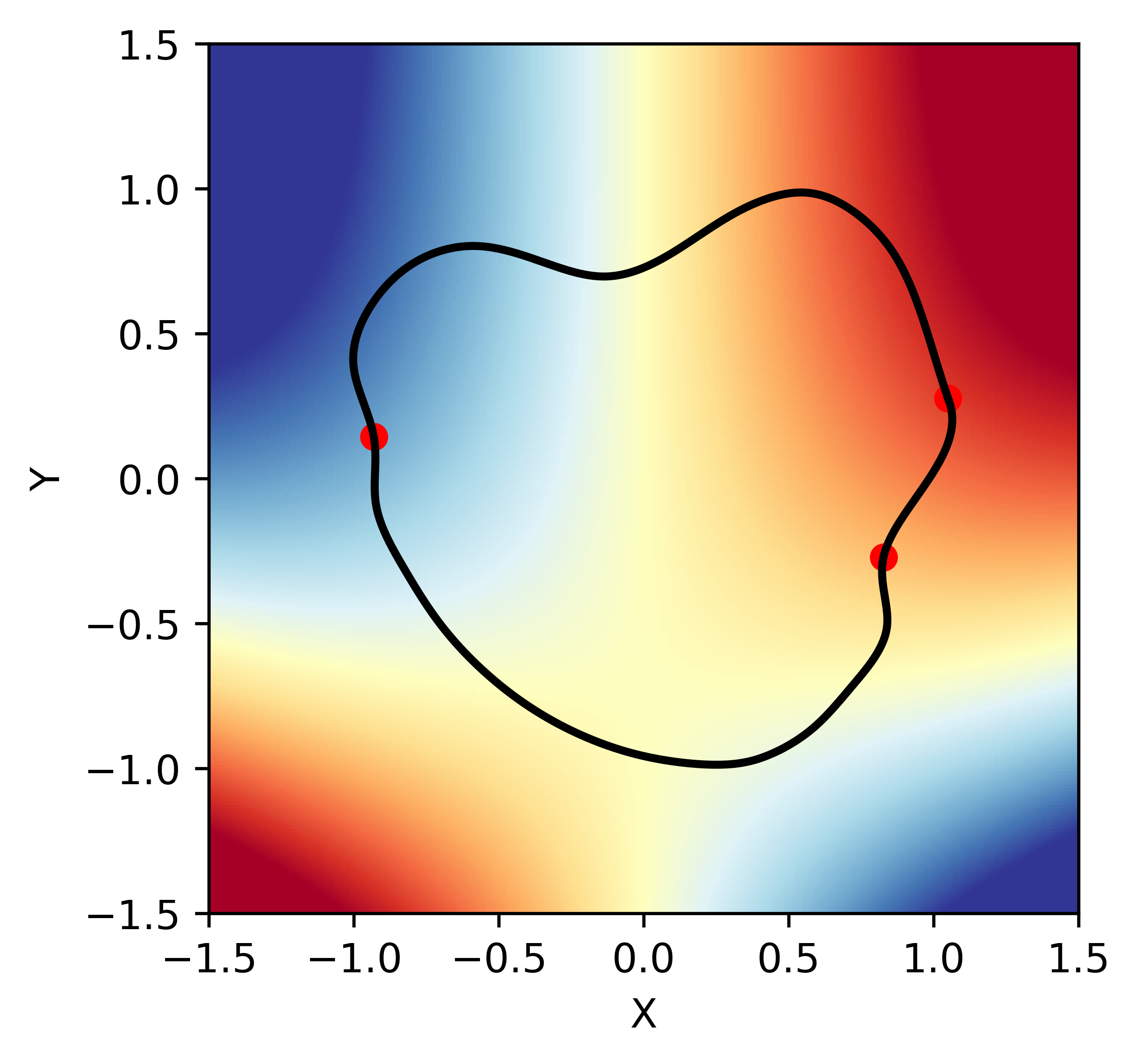}
\captionof*{figure}{Exact $u^{(1)}$.}
\end{minipage}
\begin{minipage}[t]{0.22\textwidth}
\centering
\includegraphics[width=0.9\textwidth]{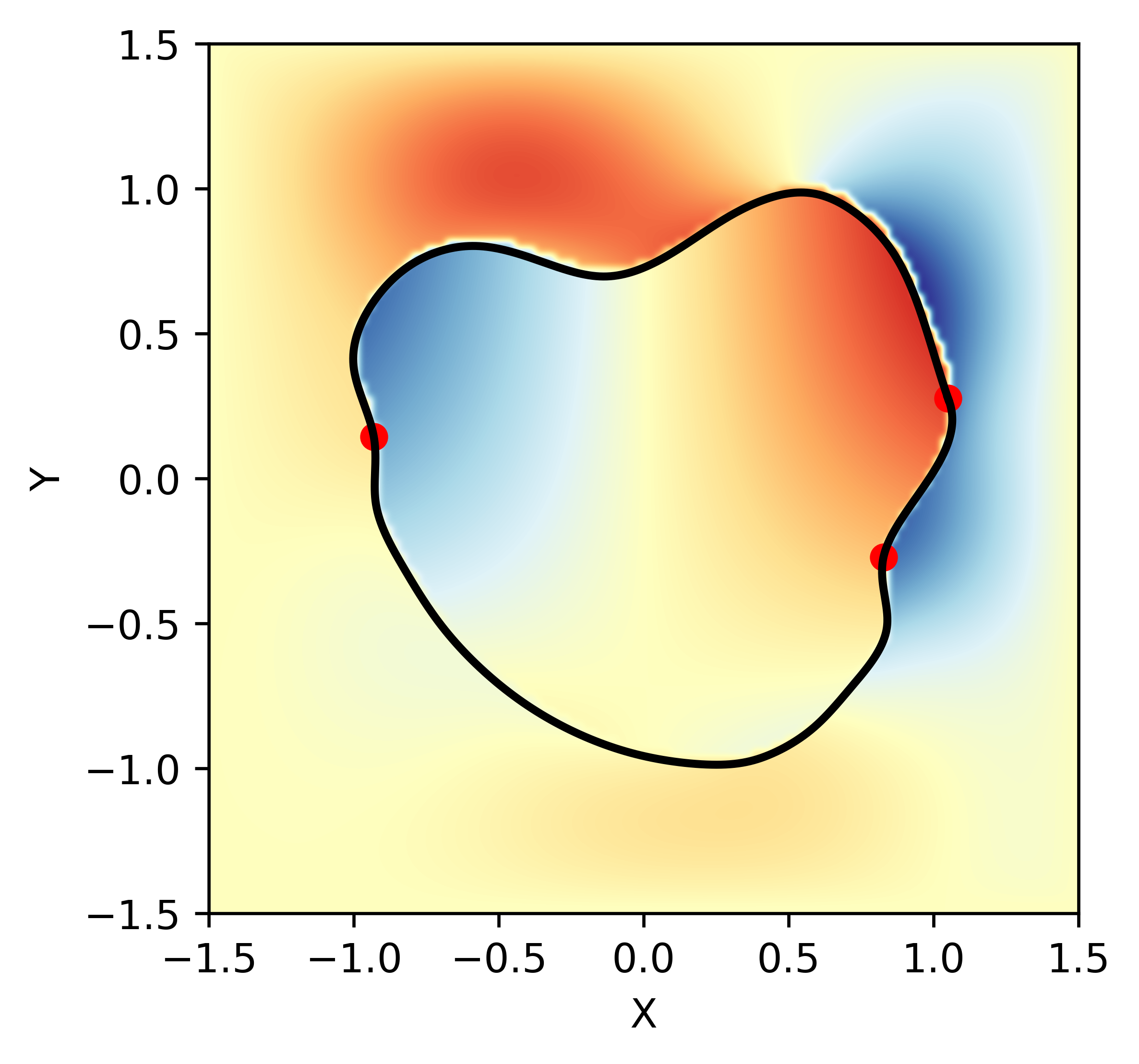}
\captionof*{figure}{Numerical $u^{(1)}$.}
\end{minipage}
\begin{minipage}[t]{0.22\textwidth}
\centering
\includegraphics[width=0.9\textwidth]{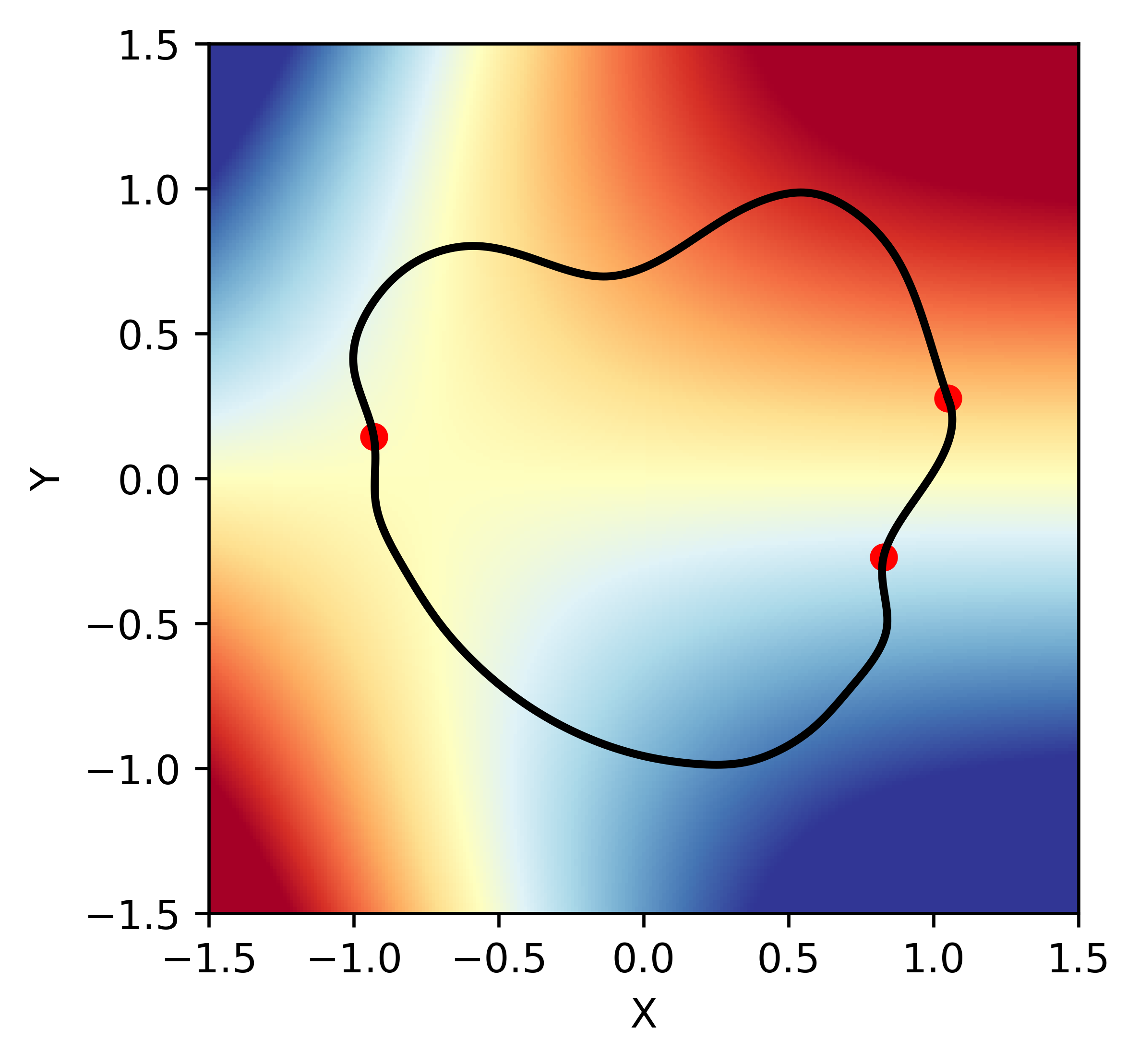}
\captionof*{figure}{Exact $u^{(2)}$.}
\end{minipage}
\begin{minipage}[t]{0.257\textwidth}
\centering
\includegraphics[width=0.9\textwidth]{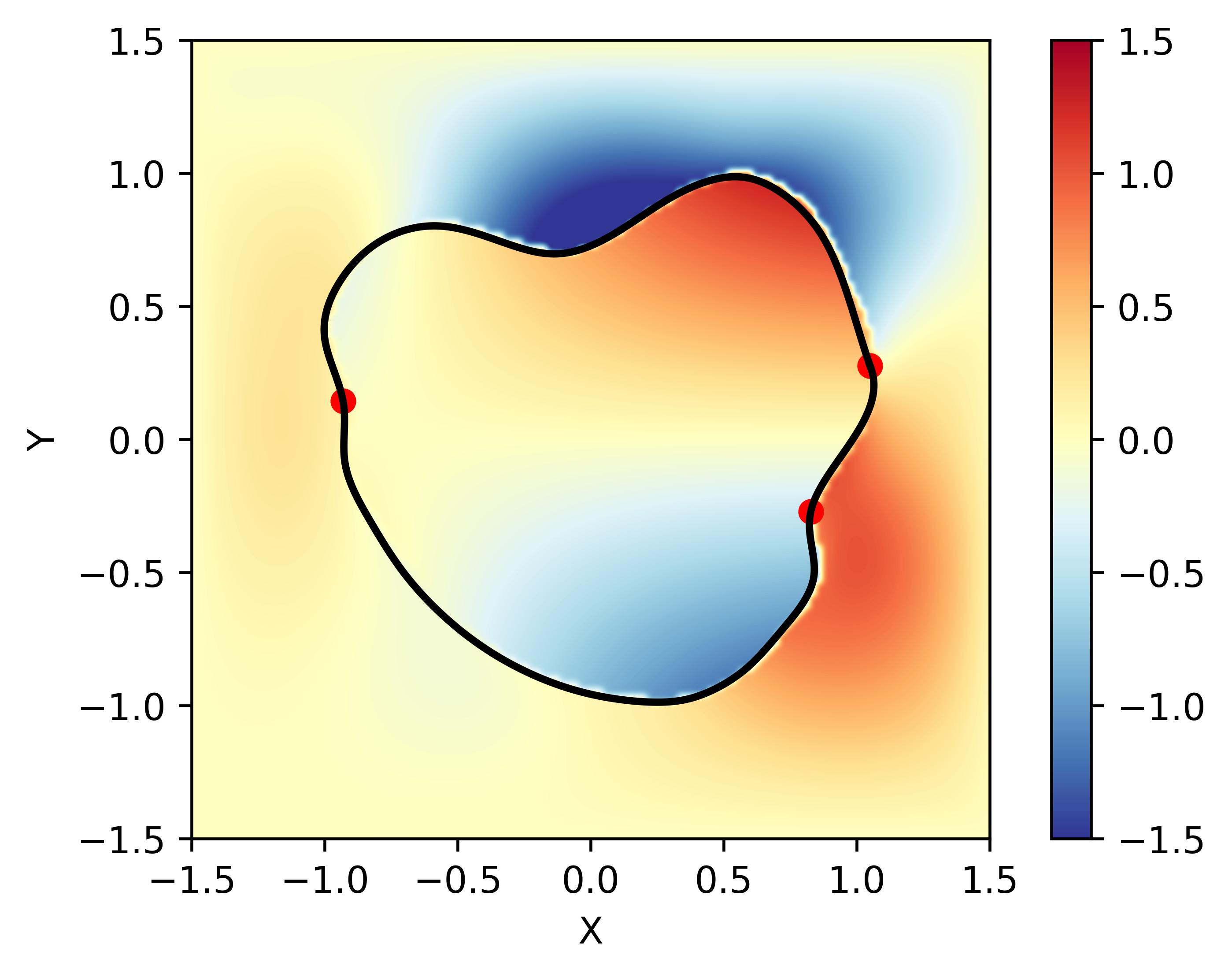}
\captionof*{figure}{Numerical $u^{(2)}$.}
\end{minipage}
\caption{Exact solution and numerical solution of the 2D Naiver equation 2 given by Strategy 1 in grid $128 \times 128$. Note that the figures show the result given by solving the corresponding interface problem in KFBI whose interior solution is the desired result for the original PDE.}
\label{fig:new_naiver}
\end{figure}

Table \ref{label_new_Naiver} presents the results for Naiver Equation 2, featuring exact solution $u^{(1)} = \sin x \cos y \\ + xy$ and $u^{(2)} = \cos x \sin y + xy$, applied to the `quasi-heart-shaped' domain with perturbed boundary with various $(\epsilon_1, \epsilon_{14}, \epsilon_{27})$ parameter values. As an example, the numerical solution obtained through Strategy 1, employing a grid size of $128 \times 128$ and perturbations $(\epsilon_1, \epsilon_{14}, \epsilon_{27}) = (+0.05, +0.05, -0.1)$, is depicted in figure \ref{fig:new_naiver}.
\begin{table}[ht]
    \centering
    \begin{tabular}{|c|c|c|c|} \hline 
         $(\epsilon_1, \epsilon_{14}, \epsilon_{27}) $ & (+0.05, +0.05, -0.1) & (-0.04, +0.02, 0) & (+0.02, -0.01, +0.06)\\ \hline 
         $L_{\infty}$ error for Strategy 1 
         & $\begin{bmatrix}  3.7E-3\\ 5.4E-3\\ \end{bmatrix}$ 
         & $\begin{bmatrix}  3.5E-3\\ 3.1E-3\\ \end{bmatrix}$ 
         & $\begin{bmatrix}  5.1E-3\\ 3.0E-3\\ \end{bmatrix}$    \\ \hline
         $L_{\infty}$ error for KFBI 
         &  $\begin{bmatrix}  1.1E-3\\ 1.2E-3\\ \end{bmatrix}$
         &  $\begin{bmatrix}  1.1E-3\\ 1.2E-3\\ \end{bmatrix}$
         &  $\begin{bmatrix}  1.1E-3\\ 1.2E-3\\ \end{bmatrix}$  \\ \hline
         $L_{\infty}$ error for Strategy 2 
         &  $\begin{bmatrix}  1.1E-3\\ 1.2E-3\\ \end{bmatrix}$
         &  $\begin{bmatrix}  1.1E-3\\ 1.2E-3\\ \end{bmatrix}$
         &  $\begin{bmatrix}  1.1E-3\\ 1.2E-3\\ \end{bmatrix}$ \\ \hline
         Iterations (KFBI) & 36 & 35 & 35 \\ \hline 
         Iterations (Strategy 2) & 23 & 20 & 22   \\ \hline 
         
         Network inference time
         & \multirow{2}{*}{0.12\bigstrut}
         & \multirow{2}{*}{0.11\bigstrut}
         & \multirow{2}{*}{0.14\bigstrut} \\
         in iteration steps & & & \\ \hline
         
         Save iterations & 36\% & 43\% & 37\% \\ \hline
    \end{tabular}
    \caption{Result of Naiver equation 2: comparison of accuracy and efficiency in solving Naiver equation 2 using different methods. The error vector is computed by the numerical solution $(u_{\mbox{numerical}}^{(1)}, u_{\mbox{numerical}}^{(2)})$ and the exact solution $(u_{\mbox{exact}}^{(1)}, u_{\mbox{exact}}^{(2)})$.}
    \label{label_new_Naiver}
\end{table}
    
In this section, we allow for perturbations to be applied to certain control points that determine the boundaries of the domain, incorporating this information as part of the model input. Experimental results demonstrate that the model under this scenario still maintains strong capabilities. Particularly noteworthy, both the error metrics and figure \ref{fig:new_naiver} representations illustrate the commendable accuracy achieved when employing the model and Strategy 1 to solve the equations.

\section{Conclusion} \label{Conclusion}

In this study, a hybrid Kernel-Free Boundary Integral Method, integrated by KFBI method and operator learning, was rigorously examined for solving parametric partial differential equations (PDEs) in complex domains. This novel approach demonstrated significant advancements in computational efficiency and accuracy.

Key findings include the model's robust generalization capabilities, allowing for accurate predictions across various boundary conditions and parameters within the same class of equations. The integration of neural network-trained models with the framework of boundary integral method significantly accelerates the computational process, reducing traditional KFBI method iterations by approximately 50\% times while retaining its inherent second-order accuracy. Moreover, the study highlighted the flexibility of the hybrid method in dealing with a range of equations in complex domains. The method's ability to transform two-dimensional problems into one-dimensional boundary problems, coupled with its independence from Green's functions, positioned it as a highly efficient computational tool.

However, challenges may appear related to generating sufficient data for training and potential issues with the size of networks for more complicated problems, which may potentially affecting inference speeds. These areas, identified for future exploration, underscore the necessity for continuous optimization of the method. To further align the hybrid KFBI Method with engineering applications, it is proposed that its scope be expanded to encompass time-dependent PDEs such as the Schrödinger, Navier-Stokes, and Maxwell equations. Additionally, the development of a three-dimensional variant of the hybrid KFBI Method is identified as a pressing objective. This expansion and enhancement are anticipated to significantly broaden the method's applicability and efficacy in complex engineering scenarios.

In conclusion, the hybrid KFBI Method, augmented with deep learning, presents a powerful tool for operator learning and solving PDEs in complex domains, offering significant improvements in computational efficiency especially under the situation of solving abundant PDEs of the same class. Its potential applications span across various fields, necessitating further research to harness its capabilities fully.

\section*{Acknowledgments}
This work is financially supported by the Strategic Priority Research Program of Chinese Academy of Sciences(Grant No. XDA25010405). It is also partially  supported by the National Key R\&D Program of China, Project Number 2020YFA0712000, the National Natural Science Foundation of China (Grant No. DMS-11771290) and the Science Challenge Project of China (Grant No. TZ2016002). Additionally, it is supported by the Fundamental Research Funds for the Central Universities. In addition, we are profoundly grateful to Professor Zhi-Qin John Xu for his unwavering guidance and unwavering support throughout our research journey. His profound expertise and invaluable insights have not only shaped the trajectory of this work but have also inspired us to reach new heights in our academic pursuits.

\newpage
\bibliography{ref}
\bibliographystyle{plain}

\newpage

\appendix
\section{Introduction to KFBI Method} \label{introduce_kfbi}
In the BI method, Green's function's explicit expression is a prerequisite, yet its exact form depends on the equation's specifics and the integration region's geometry. Despite attempts to substitute Green's function with a neural network \cite{lin2023bi}, this approach has yet to attain high accuracy levels and struggles with variable coefficients. The KFBI method \cite{ying2007kernel} circumvents the need for explicitly expressing Green's function. Here, integrals in $\eqref{one_GPU:double}$-$\eqref{one_GPU:volume}$ are treated as equivalent interface problems, facilitating the implementation of iterative steps \eqref{one_GPU:richardson1}-\eqref{one_GPU:richardson2}. Specifically, the double layer boundary integral $(W\varphi)(\mathbf{x})$ and volume integral $(Yf)(\mathbf{x})$ are reformulated as the solution $v(\mathbf{x})$ of this interface problem, with pertinent terms detailed in table \ref{tab:my_label}.
\begin{equation}
    \begin{array}{ll}
      \mathcal{L} v(\mathbf{x)} =\mathcal{F}(\mathbf{x}), & \mathbf{x} \text { in } \Omega \cup \Omega^{c}, \\
        {[v(\mathbf{x})]=\Phi(\mathbf{x}),} & \mathbf{x} \text { on } \Gamma, \\
        {\left[\partial_{\mathbf{n}}v(\mathbf{x})\right]=0,} & \mathbf{x}\text { on } \Gamma, \\
        v(\mathbf{x})=0, & \mathbf{x} \text { on } \partial \mathcal{B}.
    \end{array} \label{one_GPU:interface}
\end{equation}
\begin{table}[ht]
    \centering
    \begin{tabular}{c|c|c}
    \hline \text { Integral } & $\mathcal{F}$ & $\Phi$  \\
    \hline
    $W\varphi$ & $\mathcal{F} = 0$ & $\Phi=\varphi$  \\
    $Yf$ & $\mathcal{F}= \tilde{f}(\mathbf{x}) = \begin{cases}f(\mathbf{x}) & \text { in } \Omega \\ 0 & \text { in } \Omega^c\end{cases}$ & $\Phi = 0$ \\
\hline
\end{tabular} 
\caption{The relationship between boundary integral or volume integral and the interface problem with specific terms.} \label{tab:my_label}
\end{table}

In equation \eqref{one_GPU:interface}, the terms [$v(\mathbf{x})$] and [$\partial_{\mathbf{n}}v(\mathbf{x})$] denote the jumps in the unknown $[v(\mathbf{x})] = v^{+}(\mathbf{x}) - v^{-}(\mathbf{x})$ and its normal derivatives $[\partial_{\mathbf{n}}v(\mathbf{x})] = \partial_{\mathbf{n}}v^{+}(\mathbf{x}) - \partial_{\mathbf{n}}v^{-}(\mathbf{x})$, respectively. The function $\tilde{f}(\mathbf{x})$ represents the zero extension of the given function $f(\mathbf{x})$. Under addressing the interface problem $\eqref{one_GPU:interface}$, the computation of integrals for iterative processes and the solution $\eqref{One_GPU:fredholm_dirichlet}$ is contingent upon the numerical solution of \eqref{one_GPU:PDE} \eqref{one_GPU:PDE_D_BC} within $\Omega$. The solver for this interface problem, developed by Ying  and not the primary focus of this paper, is detailed in the literature\cite{ying2007kernel}. 

\section{The KFBI Method for Elliptic PDEs with Neumann Boundary Conditions} \label{appen::neumann::condition}
The BI method and KFBI method for elliptic PDEs with Dirichlet BVP has been introduced in Section \ref{BIM} and Appendix \ref{introduce_kfbi}. In this section, the elliptic equation $\eqref{one_GPU:PDE}$ subject to Neumann boundary condition is considered as:
\begin{equation}
    \partial_{\mathbf{n}}u(\mathbf{x}) = g_{N}(\mathbf{x}). \label{Neumann}
\end{equation}
Similarly to the Dirichlet boundary condition, the single layer potential is defined as
\begin{equation}
    (S\psi)(\mathbf{x}) := \int_{\Gamma} G(\mathbf{y}, \mathbf{x}) \psi(\mathbf{y}) d s_{\mathbf{y}} ~\text { for } ~\mathbf{x} \in \Omega \cup \Omega^{c}. \label{single}
\end{equation}

Owing to detonations defined above, the BIEs for $\eqref{one_GPU:PDE}$ and $\eqref{Neumann}$ also can be reformulated as a Fredholm boundary integral equation of the second kind\cite{kress1989linear,hsiao2008boundary}, which follows
\begin{equation}
    \frac{1}{2}\psi(\mathbf{x}) -\partial_{\mathbf{n}} (S\psi)(\mathbf{x}) + \partial_{\mathbf{n}} (Yf)(\mathbf{x}) = g_{N}(\mathbf{x}).
    \label{fredholm:neumann}
\end{equation}
The solution $u(\mathbf{x})$ to Neumann BVP $\eqref{one_GPU:PDE}$ and $\eqref{Neumann}$ is given by 
\begin{equation}
    u(\mathbf{x}) = (Yf)(\mathbf{x}) - (S\psi)(\mathbf{x}), \quad x \in \Omega.
\end{equation}
Numerically, the boundary integral equation $\eqref{fredholm:neumann}$ can be solved by simply iteration: given the artificial initial guess $\psi_{0}(\mathbf{x})$, for $k = 0, 1, 2, \cdots$, do as follows:
\begin{align}
    \partial_{\mathbf{n}} u_{k}^{+}(\mathbf{x}_{m}) = \frac{1}{2}\psi(\mathbf{x}_{m}) - \partial_{\mathbf{n}}(S\psi)(\mathbf{x}_{m}), & \quad m = 1, 2, \cdots, M,\label{neumann:richardson1} \\
    \psi_{k+1}(\mathbf{x}_{m}) = \psi_{k}(\mathbf{x}_{m}) + \gamma[\hat{g}_{N}(\mathbf{x}_{m}) - \partial_{\mathbf{n}} u_{k}^{+}(\mathbf{x}_{m})], & \quad m = 1, 2, \cdots, M, \label{neumann:richardson2}
\end{align}
where $\{\mathbf{x}_i\}_{i = 1}^M$ are control points on boundary $\Gamma$ and $\hat{g}_{N}(\mathbf{x}_{m}) := g_{N}(\mathbf{x}_{m})-\partial_{\mathbf{n}}(Yf)(\mathbf{x}_{m}),\text{ for } \mathbf{x}_{m} \in \Gamma$. Suppose $w(\mathbf{x})$ is an arbitrary piecewise smooth function with derivative  discontinuities on the interface $\Gamma$: 
\begin{equation}
    \partial_{\mathbf{n}} w^{+}(\mathbf{x}) = \lim_{z \to x, z \in \Omega} \partial _{\mathbf{n}} w(\mathbf{z}),
\end{equation}
$\partial_{\mathbf{n}} w^{-}(\mathbf{x})$ can be defined in the same way. 

As for Neumann BVP, the single layer boundary integral $\eqref{single}$ can be considered as a solution to the following interface problem:
\begin{equation}
\begin{array}{ll}
\mathcal{L} v(\mathbf{x})=0, &\text { for } \mathbf{x} \in \Omega \cup \Omega^{c},\\
{[v(\mathbf{x})]=0}, & \text { for } \mathbf{x} \in \Gamma,\\
{\left[\partial_{\mathbf{n}} v(\mathbf{x})\right]=\psi(\mathbf{x})}, & \text { for } \mathbf{x} \in \Gamma,\\
v(\mathbf{x})=0. & \text { for } \mathbf{x} \in \partial \mathcal{B}.
\end{array}
\label{single:interface}
\end{equation}

The KFBI method is distinguished by its conversion of  integral $\eqref{single}$ into the resolution of the interface problem $\eqref{single:interface}$. 
\section{Record for Control Points} \label{Record for Control Points}

\subsection{Control Points for Dumbbell-Shaped Domain} \label{appen:Control Points for Dumbbel-Shaped Domain}
The control points for the dumbbell-shaped domain in section \ref{Modified Helmholtz Equations} are given as following:

\{0.000e+00, 8.333e-02\}, 
\{-1.944e-01, 1.698e-01\}, 
\{-3.889e-01, 3.302e-01\}, 

\{-5.833e-01, 4.167e-01\}, 
\{-7.917e-01, 3.608e-01\},
\{-9.442e-01, 2.083e-01\}, 
  
\{-1.000e+00, 0.000e+00\},
\{-9.442e-01, - 2.083e-01\},
\{-7.917e-01, - 3.608e-01\},
  
\{-5.833e-01, - 4.167e-01\},
\{-3.889e-01, - 3.302e-01\}, 
\{-1.944e-01, - 1.698e-01\}, 
  
\{0.000e+00, - 8.333e-02\},
\{1.944e-01, - 1.698e-01\},
\{3.889e-01, - 3.302e-01\},
  
\{5.833e-01, - 4.167e-01\}, 
\{7.917e-01, - 3.608e-01\}, 
\{9.442e-01, - 2.083e-01\},
  
\{1.000e+00, 0.000e+00\}, 
\{9.442e-01, 2.083e-01\},
\{7.917e-01, 3.608e-01\}, 

\{5.833e-01, 4.167e-01\},
\{3.889e-01, 3.302e-01\},
\{1.944e-01, 1.698e-01\}.

\subsection{Control Points for Heart-Shaped Domain} \label{appen:Control Points for Heart-Shaped Domain}
The control points for the heart-shaped domain in section \ref{Naiver Equations} are given as following:

\{1.000e+00, 2.763e-01\},
\{9.639e-01, 5.482e-01\},
\{8.613e-01, 7.787e-01\},

\{7.076e-01, 9.328e-01\},
\{5.263e-01, 9.868e-01\}, 
\{3.070e-01, 9.118e-01\},

\{8.772e-02, 7.724e-01\},
\{-1.316e-01, 6.974e-01\},
\{-3.553e-01, 7.500e-01\},

\{-5.789e-01, 8.026e-01\},
\{-7.895e-01, 7.462e-01\},
\{-9.436e-01, 5.921e-01\},

\{-1.000e+00, 3.816e-01\}, 
\{-9.808e-01, 1.440e-01\}, 
\{-9.238e-01, - 8.645e-02\}, 

\{-8.308e-01, - 3.026e-01\},
\{-7.045e-01, - 4.980e-01\}, 
\{-5.488e-01, - 6.667e-01\},

\{-3.684e-01, - 8.035e-01\}, 
\{-1.689e-01, - 9.043e-01\},
\{4.381e-02, - 9.661e-01\},

\{2.632e-01, - 9.868e-01\}, 
\{4.271e-01, - 9.552e-01\}, 
\{5.829e-01, - 8.618e-01\},

\{7.226e-01, - 7.113e-01\},
\{8.392e-01, - 5.113e-01\},
\{9.270e-01, - 2.717e-01\}, 

\{9.815e-01, - 4.763e-03\}.


\end{document}